\documentclass[journal]{IEEEtran}  

\IEEEoverridecommandlockouts                              

\usepackage{cite}
\usepackage{amsmath,amssymb,amsfonts}
\usepackage{algorithmic}
\usepackage{graphicx}
\usepackage{float}
\usepackage{textcomp}

\usepackage[usenames,dvipsnames,svgnames,table]{xcolor}
\def\BibTeX{{\rm B\kern-.05em{\sc i\kern-.025em b}\kern-.08em
    T\kern-.1667em\lower.7ex\hbox{E}\kern-.125emX}}

\newcommand{\chg}[1]{{\textcolor{black}{ #1}}}
\newcommand{\chgcomment}[1]{{\textcolor{black}{ \unskip}}}

\definecolor{mustard}{rgb}{1.0, 0.86, 0.35}

\usepackage{framed}
\definecolor{shadecolor}{RGB}{180,180,180}

\usepackage{array}
\usepackage{gensymb}
\usepackage[utf8]{inputenc}
\usepackage{subcaption}
\usepackage{comment}
\usepackage{balance}

\usepackage[colorlinks=true,pdfpagemode=UseNone,citecolor=black,linkcolor=black,urlcolor=BrickRed]{hyperref}
\usepackage{cleveref}

\newcommand{\numlinks}{(n+1)}
\newcommand{\Q}{{\cal Q}}
\newcommand{\X}{{\cal X}} 

\newcommand{\impactmap}{\varDelta}
\newcommand{\impactmappos}{\varDelta_{q}}
\newcommand{\impactmapvel}{\varDelta_{\dot q}}

\newcommand{\poincaresection}{\mathcal{S}}
\newcommand{\pswingfootv}{p^{\rm z}_{\rm \bf sw}}
\newcommand{\vswingfootv}{\dot{p}^{\rm z}_{\rm \bf sw}}

\makeatletter
\renewcommand{\maketag@@@}[1]{\hbox{\m@th\normalsize\normalfont#1}}%
\makeatother

\usepackage{ulem}
\newcommand\ygout{\bgroup\markoverwith{\textcolor{black}{\rule[0.5ex]{2pt}{1.2pt}}}\ULon}
\renewcommand{\ygout}[1]{\unskip}

\title{\LARGE \bf  Zero Dynamics, Pendulum Models, and Angular Momentum \\
in Feedback Control of Bipedal Locomotion
\thanks{Funding for this work was provided in part by the Toyota Research Institute (TRI) under award number No.~02281 and in part by NSF Award No.~1808051. All opinions are those of the authors.}
}

\author{Yukai Gong and Jessy Grizzle
\thanks{The authors are with the College of Engineering and the Robotics Institute, University of Michigan, Ann Arbor, MI 48109 USA {\tt\small \{ykgong,grizzle\}}@umich.edu }
}

\begin{document}

\maketitle
\pagestyle{headings}
\pagenumbering{arabic}

\begin{abstract}
Low-dimensional models are ubiquitous in the bipedal robotics literature. On the one hand is the community of researchers that bases feedback control design on pendulum models selected to capture the center of mass dynamics of the robot during walking. On the other hand is the community that bases feedback control design on virtual constraints, which induce an exact low-dimensional model in the closed-loop system. In the first case, the low-dimensional model is valued for its physical insight and analytical tractability. In the second case, the low-dimensional model is integral to a rigorous analysis of the stability of walking gaits in the full-dimensional model of the robot. \chg{\textit{This paper seeks to clarify the commonalities and differences in the two perspectives for using low-dimensional models.}} In the process of doing so, we argue that angular momentum about the contact point is a better indicator of robot state than linear velocity. \textit{Concretely, we show that an approximate (pendulum and zero dynamics) model parameterized by angular momentum provides better predictions for foot placement on a physical robot (e.g., legs with mass) than does a related approximate model parameterized in terms of linear velocity.} We implement an associated angular-momentum-based controller on Cassie, a 3D robot, and demonstrate high agility and robustness in experiments.
\end{abstract}


\section{Introduction}
\label{sec:Intro}

Models of realistic bipedal robots tend to be high-dimensional, hybrid, nonlinear systems. This paper is concerned with two major themes in the literature for ``getting around'' the analytical and computational obstructions posed by realistic models of bipeds. 

On the one hand are the broadly used, simplified pendulum models \cite{miura1984dynamic,kajita20013d,BL89,pratt2006capture,englsberger2011bipedal,WANGChevallereau2011,xiong2019orbit} that provide a computationally attractive model for the center of mass dynamics of a robot. When used for control design, the fact that they ignore the remaining dynamics of the robot generally makes it impossible to prove stability properties of the closed-loop system. Despite the lack of analytical backing, the resulting controllers often work in practice when the center of mass is well regulated to match the assumptions underlying the model. Within this context, the dominant low-dimensional pendulum model by far is the so-called linear inverted pendulum model, or LIP model for short, which captures the  center of mass dynamics of a real robot correctly when, throughout a step, the following conditions hold: (i) the center of mass (CoM) moves in a straight line; and (ii), the robot's angular momentum about the center of mass ($L_c$) is zero (or constant). \chg{This latter condition can be met by designing a robot to have light legs, such as the Cassie robot by Agility Robotics \cite{ARsite2021}, or by deliberately regulating $L_c$ to zero \cite{GOSWAMIA04,1442139}. When $L_c$ cannot be regulated to a small value, an MPC feedback control law based on the LIP model has been proposed to minimize zero moment point (ZMP) tracking error and CoM jerk \cite{KaKaKaFuHaYoHi03,nishiwaki2006high, wieber2006trajectory}. The effects of $L_c$ can be compensated with ZMP, making a real robot's CoM dynamics the same as those of a LIP \cite{englsberger2012integration}. Alternatively, $L_c$ can be approximately predicted and used for planning \cite{seyde2018inclusion,lee2007reaction}. }

On the other hand, the control-centric approach called the Hybrid Zero Dynamics provides a mathematically-rigorous gait design and stabilization method for realistic bipedal models \cite{WGCCM07,yang2009framework,Martin2014Design,Zhao2015hybrid,reher2016realizing,ReCoHeHuAm16,griffin2015nonholonomic,Agrawal2017First,gurriet2018towards,da2019combining} without restrictions on robot or gait design. In this approach, the links/joints of the robot are synchronized via the imposition of ``virtual constraints'', meaning the constraints are achieved through the action of a feedback controller instead of contact forces. As opposed to physical constraints, virtual constraints can be re-programmed on the fly. Like physical constraints, imposing a set of virtual constraints results in a reduced-dimensional model. The term ``zero dynamics'' for this reduced dynamics comes from the original work of \cite{ISI95,BYRNESC91}. The term ``hybrid zero dynamics'' or HZD comes from the extension of zero dynamics to (hybrid) robot models in \cite{WEGR02}. A downside of this approach, however, has been that it lacked the ``analytical tractability'' provided by the pendulum models\footnote{The approaches in \cite{poulakakis2009spring,hereid2014dynamic,chen2020optimal} to build reduced-order models via embeddings is a step toward attaching physical significance to the zero dynamic models.}, and it requires non-trivial time to find optimal virtual constraints for a realistic model.

\begin{figure}[b]
    \centering
    \includegraphics[width=0.47\textwidth]{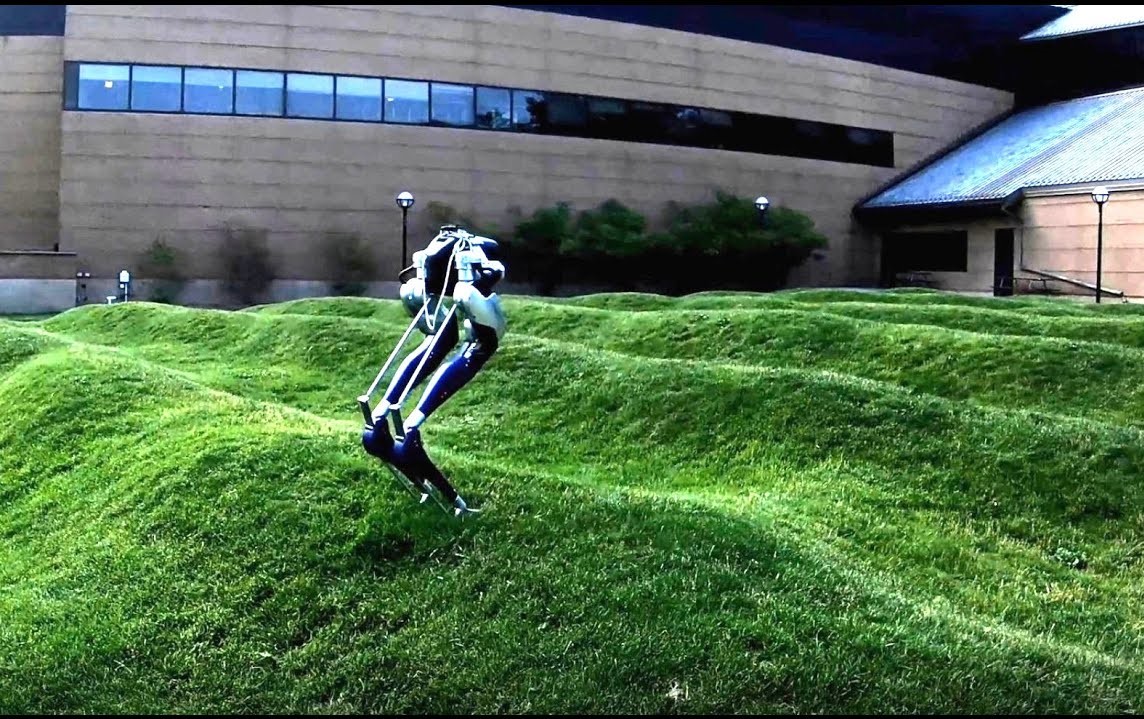}
    \caption{Cassie Blue, by Agility Robotics, on the iconic University of Michigan Wave Field.}
    \label{fig:Cassie_WaveField}
\end{figure}

\chg{While CoM velocity is the most widely used variable ``to summarize the state'' of a bipedal robot, angular momentum about the contact point has also been valued by multiple researchers. In \cite{SAFU90}, angular momentum is chosen to represent a biped's state and it is regulated by stance ankle torque. In \cite{WEGRKO03}, the  relative degree three property of angular momentum motivated its use as a state variable in the zero dynamics. In \cite{grizzle2005nonlinear,azad2013angular}, control laws for robots with an unactuated contact point were proposed to exponentially stabilize them about an equilibrium. In \cite{griffin2015nonholonomic}, angular momentum is explicitly used for designing nonholonomic virtual constraint. In \cite{powell2016mechanics}, angular momentum is combined with the LIP model to yield a controller that stabilizes the transfer of angular momentum from
one leg to the next through continuous-time (single-support-phase) control coupled
with a hybrid model that captures impacts that occur at
foot strike. In \cite{gong2021one}, the accuracy of the angular-momentum-based LIP model during the continuous phase is emphasized; as opposed to \cite{powell2016mechanics}, angular momentum is allowed to passively evolve according to gravity during each single support phase, and foot placement is used to regulate the estimated angular momentum at the end of the ensuing step.}

\ygout{In \mbox{\cite{de2019essential}}, De-Le{\'o}n-G{\'o}mez et al. introduced the concept of the \textit{essential dynamics} of a fully actuated robot and studied its relations with the 3D LIP model. The essential dynamics is parameterized by the evolution of the ``non-controlled (or free) internal state'' of the robot and the desired position of the ZMP. Similar to the zero dynamics, it provides an exact representation of the robot's ``centroidal dynamics'' and its calculation as a function of the robot's body coordinates is the same as that of the zero dynamics. The main differences with respect to the hybrid zero dynamics are that (1) the essential dynamics is not tied to the zeroing of an output (making it more general, in a certain sense) and (2) a formal stability theory has not yet been developed for it (leaving work to be done).}

\subsection{Objectives}


The objectives of this paper are two-fold. Firstly, we seek to contribute insight on how pendulum models relate among one another and to the dynamics of a physical robot. We demonstrate that even when two pendulum models originate from the same (correct) dynamical principles, the approximations made in different coordinate representations lead to non-equivalent approximations of the dynamics of a (realistic) bipedal robot. Secondly, we seek a rapprochement of the most common pendulum models and the hybrid zero dynamics of a bipedal robot. Both of these objectives are addressed for planar robot models. \chg{The extension to 3D is not attempted here, primarily to keep the arguments as transparent as possible.} \ygout{While we suspect that a 3D extension is possible along the lines of the 3D essential dynamics \mbox{\cite{de2019essential}} or the VLIP \mbox{\cite{luo2019walking}} (LIP with vertical oscillations), it is not attempted here, primarily to keep the arguments as transparent as possible; we will make use of the essential dynamics in 2D.}

The first point that, approximations made in different coordinate representations lead to non-equivalent approximations of the dynamics of a real robot, is important in practice; hence we elaborate a bit more here, with details given in Sect.~\ref{sec:ControlZDoffzeroDynamics}. Let's only consider trajectories of a robot where the center of mass height is constant, and therefore, the velocity and acceleration of the center of mass height are both zero. In a realistic robot, the angular momentum about the center of mass, denoted by $L_c$, contributes to the longitudinal evolution of the center of mass, though it is routinely dropped in the most commonly used pendulum models. Can dropping $L_c$ have a larger effect in one simplified model than another? 

In the standard 2D LIP model, the coordinates are taken as the horizontal position and velocity of the center of mass \textit{and the time derivative of $L_c$ is dropped from the differential equation for the velocity}. It follows that the term being dropped is a high-pass filtered version of $L_c$, due to the derivative. Moreover, the derivative of $L_c$ is directly affected by the motor torques, which are typically ``noisy'' (have high variance) in a realistic robot. On the other hand, in a less frequently used representation of a 2D inverted pendulum \cite{griffin2015nonholonomic,GriffinIJRR2016,powell2016mechanics,gong2021one}, the coordinates are taken as angular momentum about the contact point and the horizontal position of the center of mass, \textit{and $L_c$ is dropped from the differential equation for the position}. In this model, $L_c$ (and not $\dot{L}_c$) shows up in the second derivative of the angular momentum about the contact point. It follows that variations in $L_c$ are low-pass filtered in the second representation as opposed to high-pass filtered in the first, and thus, speaking intuitively, neglecting $L_c$ should induce less approximation error in the second model. More quantitative results are shown in the main body of the paper.


\chg{In this paper, we focus on the underactuated single support phase dynamics and assume an instantaneous double support phase. The reader is referred to existing  literature on how pendulum models \cite{kajita2003biped,xiong2021global} and Hybrid Zero Dyanmics \cite{hamed2011stabilization,ma2019first,reher2020algorithmic} handle non-instantaneous double support phases; the topic is not discussed in this paper. We provide models for a robot with non-trivial feet. While most of the results are demonstrated for robots with point feet, we briefly show that the conclusions we obtained for robots with point feet still apply to robots with non-trivial feet.} \chg{Further studies of how pendulum models and Hybrid Zero Dynamics handle non-trivial feet can be found in \cite{KaKaKaFuHaYoHi03, 8460572, CHOIJ05a, paredes2020dynamic}}


\ygout{Brief Remarks on the Literature}

\ygout{While maintaining ``balance'' is a critical problem in bipedal locomotion, how the notion of ``balance'' is quantified and used in the control design can be very different.  A common approach is to summarize the status of a nonlinear high-dimensional robot model with a few key variables.
The most frequently proposed variables as surrogates for ``balance'' include Center of Mass (COM) Velocity \mbox{\cite{kajita20013d,da20162d,HartleyGrizzleCCTA2017,Haribexo2018,Yukai2018,xiong2019orbit,takenaka2009real,pratt1999exploiting,rezazadeh2015spring,castillo2021robust,xie2020learning}}, Capture Point \mbox{\cite{pratt2006capture,englsberger2011bipedal}}, Zero Moment Point \mbox{\cite{1241826,7363473}}, and Angular Momentum \mbox{\cite{griffin2015nonholonomic,powell2016mechanics,GriffinIJRR2016,dai2021bipedal}}.}

\ygout{This paper is mostly closely related to works that equate ``balance'' with an asymptotically stable periodic orbit. Within this category, we would include all papers on self-synchronization and self-stabilization \mbox{\cite{razavi2017symmetry,luo2018self,luo2018self}}, most papers on the compass gait biped \mbox{\cite{goswami1997limit,hiskens2001stability,pekarek2007discrete,znegui2020design}}, many papers on simplified pendulum models as found in the survey \mbox{\cite{he2017survey}}, and papers that treat virtual constraints or zero dynamics \mbox{\cite{GRAB01,WGCCM07,chevallereau2003time,yang2009framework,WANGChevallereau2011,Ames2012Dynamically,SrPaGr2012,MaPoSc14,embry2016unified,da2019combining}}}.

\subsection{Summary of Main Contributions}
This paper makes the following contributions to pendulum models and zero dynamics:
\begin{itemize}
        \item Starting from a common physically correct set of equations for real robots, we sequentially enumerate the approximations made to arrive at various reduced order pendulum models. \chg{This is done for bipedal robots with and without ankle torque.}
    \item Several advantages of using angular momentum about the contact point as a key variable to summarize the dynamics of a bipedal robot are discussed and demonstrated.
    \item We show that a pendulum model parameterized by center of mass horizontal position and angular momentum about the contact point provides a higher fidelity representation of a physical robot than does the standard LIP model, which is a pendulum model parameterized by center of mass horizontal position and velocity. We identify the source of improvement as how the angular momentum about the center of mass is treated in the two models.
    \item \chg{For a set of non-holonomic virtual constraints, we derive the zero dynamics in a set of coordinates compatible with an angular-momentum-based pendulum model. We provide a precise sense in which the pendulum model is an approximation to the zero dynamics and demonstrate what this means for closed-loop stability of the full-order robot model. }
    \item We formulate a foot placement strategy based on a high fidelity, one-step-ahead prediction of angular momentum about the contact point.
    \item We demonstrate that the resulting controller achieves highly dynamic gaits on a Cassie-series bipedal robot.
\end{itemize}
The last two contributions were introduced in \cite{gong2021one} and are used here to demonstrate the utility of the proposed results.
We will use both Rabbit \cite{CHABAOPLWECAGR02} and Cassie Blue, shown in Fig.~\ref{fig:Cassie_WaveField}, to illustrate the developments in the paper. Experiments will be conducted exclusively on Cassie. 

\begin{figure} 
    \centering
    \includegraphics[width=0.5\textwidth]{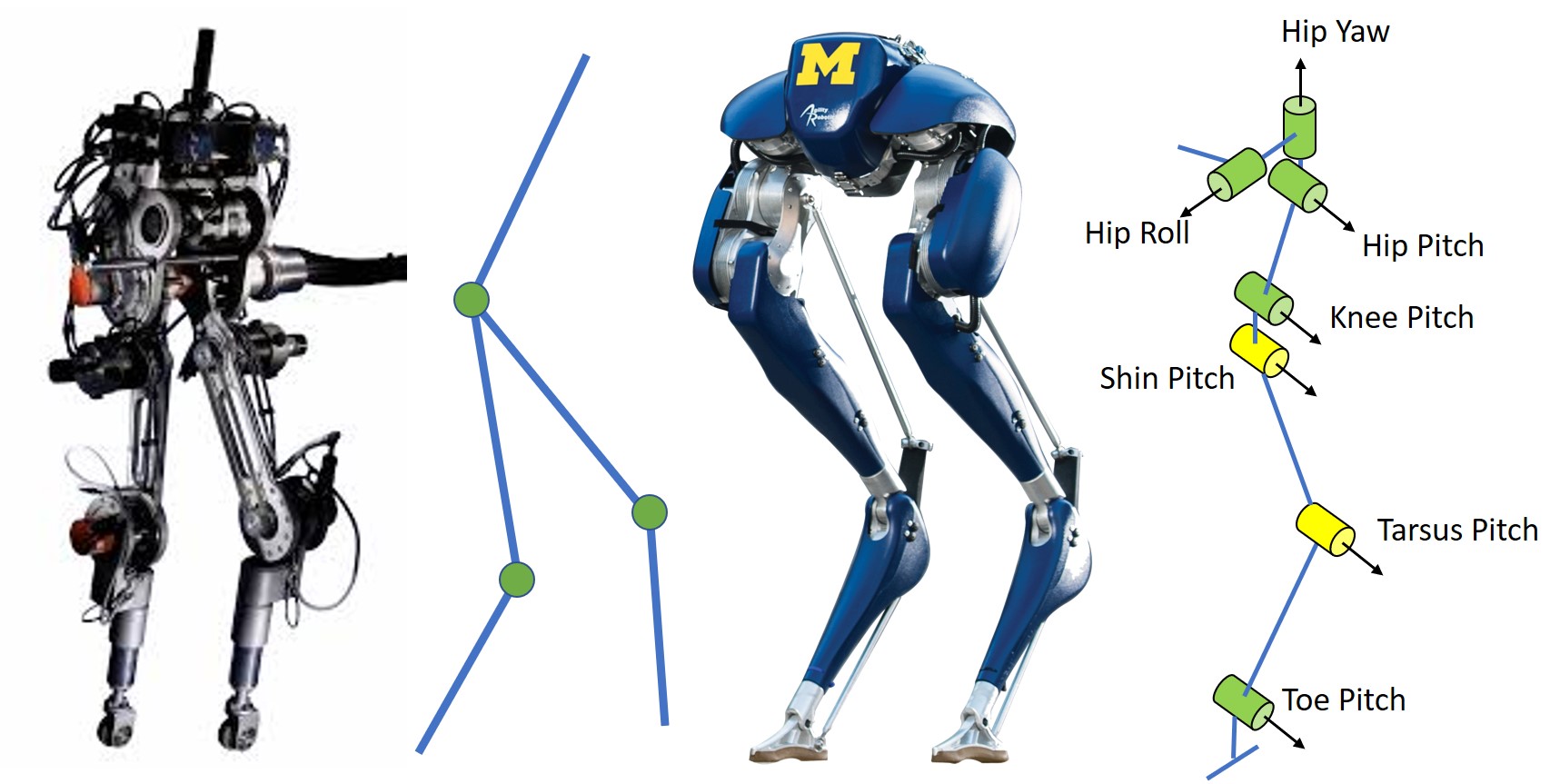}
    \caption{Rabbit and Cassie. Rabbit is planar robot with 2 joints on each leg and Cassie is 3D robot with 7 joints on each leg.}
    \label{fig:Rabbit_Cassie}
\end{figure}

Rabbit is a 2D biped with five links, four actuated joints, and a mass of 32~Kg; see Fig.~\ref{fig:Rabbit_Cassie}. Each leg weighs 10 kg, with 6.8 kg on each thigh and 3.2 kg on each shin. The Cassie robot designed and built by Agility Robotics weighs 32~Kg. It has 7 deg of freedom on each leg, 5 of which are actuated by motors and 2 are constrained by springs; see Fig. \ref{fig:Rabbit_Cassie}. A floating base model of Cassie has 20 degrees of freedom. Each foot of the robot is blade-shaped and provides 5 holonomic constraints when firmly in contact with the  ground. Though each of Cassie's legs has approximately 10 kg of mass, most of the mass is concentrated on the upper part of the leg. In this regard, the mass distribution of Rabbit is more typical of current bipedal robots seen in labs, which is why we include the Rabbit model in the paper.

\subsection{Organization} 

\chg{The remainder of the paper is organized as follows. Sect.~ \ref{Sec:HyrbidRobotModel} introduces the full hybrid model of a bipedal robot with instantaneous double support phases. It also provides the robot's center of mass dynamics in different coordinates. Sect.~ \ref{sec:AngularMomentumGeneral} summarizes several desirable properties of angular momentum about contact point. Sect.~ \ref{sec:PendulumModels} goes through the approximations that must be made to the center of mass dynamics in order to arrive at the most popular pendulum models. The approximation errors of the resulting pendulum models are analyzed and compared. Sect.~ \ref{Sec:Controller} uses a one-step-ahead prediction of angular momentum to decide foot placement. This provides a feedback controller based on predicted angular momentum that will stabilize a 3D pendulum. Sect.~ \ref{sec:ControlZDoffzeroDynamics} provides background on a feedback control paradigm based on virtual constraints and the attendant zero dynamics. One of the pendulum models is shown to be a clear approximation to the zero dynamics, and the implications for stability analysis are discussed and illustrated by numerical computation. In Sect.~ \ref{sec:virtual_constraint_Cassie}, we design the virtual constraints for a Cassie-series bipedal robot. In Sect.~ \ref{Sec:Experiment}, we provide our path to implementing the controller on Cassie Blue. Additional virtual constraints are required beyond a path for the swing foot, and we provide ``an intuitive'' method for their design. Sect.~ \ref{Sec:ExperimentResults} shows the results of experiments. Conclusion are give in Sect.~ \ref{Sec:Conclusion}. }

\section{Swing Phase and Hybrid Model}
\label{Sec:HyrbidRobotModel}


This section introduces the full-dimensional swing-phase model that describes the mechanical model when the robot is supported on one leg and a hybrid representation used for walking that captures the transition of support legs. The section concludes with a summary of a few model properties that are ubiquitous when discussing low-dimensional pendulum models of walking. 

\subsection{Full-dimensional Single Support Model}
 We assume a planar bipedal robot satisfying the specific assumptions in \cite[Chap.~3.2]{WGCCM07} and \cite{WEGRKO03}, which can be summarized as a \ygout{passive} revolute point contact with the ground, no slipping, all other joints are independently actuated, and all links are rigid and have mass. The gait is assumed to consist of alternating phases of single support (one ``foot'' on the ground), separated by instantaneous double support phases (both feet in contact with the ground), with the impact between the swing leg and the ground obeying the non-compliant, algebraic contact model in \cite{HUCH92,HUR93} (see also \cite[Chap.~3.2]{WGCCM07}).
 
\chg{The contact point with the ground, which we refer to as the stance ankle, can be passive or actuated. Even when actuated, the stance ankle is ``weak'' in the sense that only limited torque can be applied before the foot rolls about one of its extremities. The swing ankle is not weak, however, because it only needs to regulate the orientation of the swing foot. To accommodate both actuation scenarios, we will routinely separate the stance ankle actuation from other actuators on the robot so that it can be either set to zero or appropriately exploited.}

We assume a world frame $(x,z)$ with the right-hand rule. We assume the swing-phase (pinned) Lagrangian model is derived in coordinates $q:=(q_0, q_1, \ldots, q_n) \in Q$, where $q_0$ is an absolute angle (referenced to the $z$-axis of the world frame) and $q_b:= (q_1, \ldots, q_n)$ are body coordinates. Furthermore, we reference the contact point (i.e., stance ankle) to the origin of the world frame. 

With the above sets of assumptions, \chg{the robot in single-support is either fully actuated or has one degree of underactuation. Moreover, $q_0$ is a cyclic variable (of the kinetic energy).} It follows that the dynamic model can be expressed in the form 
\begin{equation}
    \label{eq:SwingPhasePinnedModel}
    D(q_b)\ddot{q} +  C(q,\dot{q}) \dot{q} + G(q) = B(q) u,
\end{equation}
where the vector of motor torques $u\in \mathbb{R}^n$ and the torque distribution matrix has full column rank. The model is written in state space form by defining
\begin{equation}
\label{eqn:grizzle:modeling:model_walk}
\begin{aligned}
    \dot x = & \left[
      \begin{array}{c}
        \dot{q}\\
        D^{-1}(q_b)
        \left[-C(q,\dot{q})\dot{q} - G(q) + B(q) u\right]
      \end{array}\right]\\
   =: & f(x) + g(x) u
\end{aligned}
\end{equation}
where $x := (q;\dot{q})$. The state space of the model is $\X =
T\Q$. For each $x\in \X$, $g(x)$ is a $2 \numlinks \times
\numlinks$ matrix. In
natural coordinates $(q;\dot{q})$ for $T\Q$, $g$ is independent of
$\dot{q}$.

\subsection{Full Dimensional Hybrid Model}
In the above, we implicitly assumed left-right symmetry in the robot so that we could avoid
the use of two single-support models---one for each leg playing the
role of the stance leg---by relabeling the robot's coordinates at
impact, thereby swapping their roles. Immediately after swapping, the former swing
leg is in contact with the ground and is poised to take on the role of
the stance leg. 
The result of the
impact and the relabeling of the states provides an expression
\begin{equation}\label{eqn:grizzle:modeling:delta}
  x^+ = \impactmap(x^-)
\end{equation}
where $x^+:=(q^+; \dot{q}^+)$ (resp. $x^-:=(q^-; \dot{q}^-)$) is the
state value just after (resp. just before) impact and
\begin{equation}\label{eqn:grizzle:modeling:deltadefined}
  \impactmap(x^-) :=
  \left[
    \begin{array}{c}\smallskip
      \impactmappos(q^-) \\ \impactmapvel(q^-)\, \dot{q}^-
    \end{array}
  \right].
\end{equation}
A detailed derivation of the impact map is given in \cite{WGCCM07}, showing that it is linear in the generalized velocities.

A hybrid model of walking is
obtained by combining the single support model and the impact model to
form a system with impulse effects \cite{GRABPL01}. \chg{A non-instantaneous double support phase can be added \cite{hamed2011stabilization,reher2020algorithmic}, but we choose not to do so here.}  Even though the mechanical model of the robot is time-invariant, we will allow feedback controllers for \eqref{eqn:grizzle:modeling:model_walk} that are time varying. So that the hybrid model in closed loop can be analyzed with tools developed for time-invariant hybrid systems, we do the standard ``trick'' of adding time as a state variable via $\dot{\tau}=1$. The guard condition (aka switching set) for terminating a step is
\begin{equation}
  \label{eqn:grizzle:modeling:S}
  \poincaresection := \{ (q,\dot{q}) \in T\Q~|~\pswingfootv(q) = 0,\; \vswingfootv(q, \dot{q}) <0 \},
\end{equation}
where $\pswingfootv(q)$ is the vertical height of the swing foot. It is noted that $\poincaresection$ is independent of time. Combining \eqref{eqn:grizzle:modeling:model_walk}, \eqref{eqn:grizzle:modeling:delta} with the guard set and time gives the hybrid model
\begin{equation}\label{eqn:grizzle:modeling:full_hybrid_model_walking}
\Sigma: \begin{cases}
  \begin{aligned}
    \dot x & = f(x) + g(x) u   & x^- &\notin \poincaresection\\
    \dot \tau & = 1 \\
       x^+ & = \impactmap(x^-) & x^- &\in \poincaresection \\
       \tau^+ & = 0. 
  \end{aligned}
  \end{cases}
\end{equation}
It is emphasized that the guard condition for re-setting the ``hybrid time variable'', $\tau$, is determined by foot contact.


\subsection{Center of Mass Dynamics in Single Support}
\label{sec:LowDimesionDynamics}
While \eqref{eqn:grizzle:modeling:model_walk} is typically high dimensional and nonlinear, standard mechanics yields simpler equations for the evolution of the center of mass. For succinctness, we only consider the planar case and define the following variables:
\begin{itemize}
    \item $(x_{\rm c},z_{\rm c})$ : CoM position in the frame of the contact point.
        \item $v_{\rm c}$ :  CoM velocity in $x$-direction. The velocity in $z$-direction is denoted by $\dot{z}_{\rm c}$ 
    \item $L_{\rm c}$ : $y$-component of Angular momentum about CoM.
    \item $L$ : $y$-component of Angular momentum about contact point.
    \item $u_a$ : ankle torque at the contact point.
\end{itemize}
In addition, we note the following (standard) result
\begin{equation}
    \label{eq:AngularMomentumCoM}
 L = L_c + m \begin{bmatrix}
 x_c \\ z_c 
 \end{bmatrix} \wedge   \begin{bmatrix}
 \dot{x}_c \\ \dot{z}_c 
 \end{bmatrix}
\end{equation}
where $\wedge$ is the 2D version of cross product
$$\begin{bmatrix}
 x_c \\ z_c 
 \end{bmatrix} \wedge  \begin{bmatrix}
 \dot{x}_c \\ \dot{z}_c 
 \end{bmatrix} := \left(\begin{bmatrix}
 x_c \\ 0 \\z_c 
 \end{bmatrix} \times \begin{bmatrix}
 \dot{x}_c \\ 0\\ \dot{z}_c 
 \end{bmatrix}\right) \bullet \begin{bmatrix}
 0 \\ 1 \\0
 \end{bmatrix}. $$
 We refer to \eqref{eq:AngularMomentumCoM} as the \textit{angular momentum transfer formula} because it relates angular momentum determined about two different points.

In the following, we provide the CoM dynamics for two sets of coordinates 
\begin{itemize}
    \item $(x_{\rm c}, v_{\rm c})$
    \item $(x_{\rm c}, L)$, and 
    \item $(\theta_{\rm c}, L)$,
\end{itemize}
\chg{
where
\begin{equation}
\label{eq:thetac}
  \theta_c := {\rm atan}(x_c/z_c)  
\end{equation}
and we assume that $z_c>0$.}
We will subsequently dedicate Sect.~\ref{sec:PendulumModels} to establishing connections between pendulum models and zero dynamics, \chg{which will allow the zero dynamics to be intuitively grounded in physics.} 
 
 \noindent \textbf{Case 1:} $(x_c, v_c)$ \textbf{Horizontal Position and Velocity}  Differentiating \eqref{eq:AngularMomentumCoM} and using $v_c=\dot{x}_c$ results in
\begin{equation}
\label{eqn:zdOtt}
    \begin{aligned}
      \dot{x}_c&= v_c\\
      \dot{v}_c&=\frac{g}{z_c} x_c + \frac{\ddot{z}_c}{z_c}x_c -\frac{\dot{L}_c}{mz_c} +\frac{u_a}{mz_c}.
    \end{aligned}
\end{equation}
 \chg{In general, $z_c$ depends on $q$, $\dot{z}_c$ and $L_c$ depend on both $q$ and  $\dot{q}$. While $\dot{L}_c$ and $\ddot{z}_c$ depend on $q$, $\dot{q}$, and the motor torques $u$, it is more typical to replace the motor torques by the ground reaction forces. In particular, one uses $\ddot{z}= g - \frac{1}{m}F_z$ and $\dot{L}_c:=\frac{d}{dt}L_c = x_c F_z - z_c F_x + u_a$, where $F_x$ and $F_z$ are the horizontal and vertical components of the ground reaction forces. In turn, the ground reaction forces can be expressed as functions of $q$, $\dot{q}$, and the motor torques, $u$.}

 \noindent \textbf{Case 2:} $(x_c, L)$  \textbf{Angular Momentum and Horizontal Position:} Manipulating \eqref{eq:AngularMomentumCoM} and using $\dot{L}=m g x_c + u_a$ results in
\begin{equation}
    \label{eqn:ZDnaturalCoordinates}
    \begin{aligned}
       \dot{x}_c&= \frac{L}{m z_c} + \frac{\dot{z}_c}{z_c}x_c - \frac{L_c}{m z_c}\\
       \dot{L}&=m g x_c + u_a.
    \end{aligned}
\end{equation}
The remarks made above on $z_c$, $\dot{z}_c$, and $L_c$ apply here as well.

  \noindent \textbf{Case 3:} $(\theta_c, L)$ \textbf{Alternative absolute angle (cyclic variable):} \chg{Differentiating \eqref{eq:thetac} yields
            \begin{equation}
    \label{eq:thetaCdot}
  \dot{\theta}_c= \frac{L-L_c}{m r_c^2(q_b)}.
\end{equation}
}  Combining \eqref{eq:AngularMomentumCoM} and \eqref{eq:thetaCdot} yields
        \begin{equation}
    \label{eqn:JWG:ZD_theta_c}
       \begin{aligned}
       \dot{\theta}_c&= \frac{L-L_c}{m r_c^2}  \\
       \dot{L}&=m g r_c \sin(\theta_c) + u_a.
    \end{aligned}
\end{equation}
\ygout{In comparison to (an eqn removed), it's now easier to interpret the way in which the actuated states enter the dynamics, namely, through $r_c$, the length of a pendulum representation of the center of mass, and $L_c$, the angular momentum about the center of mass.}It is remarked that the derivatives of the generalized coordinates only appear through $L_c$. In the following, we will keep the discussion primarily focused on \eqref{eqn:ZDnaturalCoordinates}, but most of the results apply to \eqref{eqn:JWG:ZD_theta_c} as well; see Appendix~\ref{sec:ThetacModel}.

\section{Angular Momentum about Contact Point}
\label{sec:AngularMomentumGeneral}

In this paper, we are focusing on the angular momentum about the contact point, $L$, as a replacement for the center of mass velocity, $v_{\rm c}$, which is used as an indicator of walking status in many other papers \cite{Kajita1991,da2019combining,xiong2019orbit,Haribexo2018}. Specific to this paper, $L$ is also a state of the zero dynamics. Before we proceed to that, it is beneficial to explain why $L$ can replace $v_{\rm c}$, summarize some general properties of $L$, and highlight some of its advantages versus $v_{\rm c}$. More specific advantages of using $L$ in the zero dynamics and the LIP model will be discussed in later sections.

We first need to answer why $L$ can replace $v_{\rm c}$ as an indicator of walking. The relationship between angular momentum and \textbf{linear momentum} for a 3D bipedal robot is
\begin{equation}
\label{eq:AngularMomentumLinearMomentum}
    L = L_{\rm c} + p_{\rm c} \wedge m v_{\rm c},
\end{equation}
where $L_{\rm c}$ is the angular momentum about the center of mass, $v_{\rm c}$ is the linear velocity of the center of mass, $m$ is the total mass of the robot, and $p_{\rm c}$ is the vector emanating from the contact point to the center of mass.

 For a bipedal robot that is walking instead of doing somersaults, it is reasonable to focus on gaits where the angular momentum about the center of mass oscillate about zero (e.g., arms are not rotating as in a flywheel). \chg{The oscillating property of $L_{\rm c}$ is discussed in \cite{MALIGRRUSE10,wieber2006holonomy}.} When $L$ oscillates about zero, \eqref{eq:AngularMomentumLinearMomentum} implies that the difference between $L$ and $p_{\rm c}\wedge m v_{\rm c}$ also oscillates about zero, which we will write as
 \begin{equation}
\label{eq:AngularMomentumLinearMomentum02}
  L - p_{\rm c} \wedge mv_{\rm c}  = L _{\rm c} \text{~oscillates about~} 0.
\end{equation}
From \eqref{eq:AngularMomentumLinearMomentum02}, we see that we approximately obtain a desired linear velocity by regulating $L$. Hence, \chg{in walking robots without a flywheel,} one can replace the control of linear velocity with control of angular momentum about the contact point.

\begin{figure*}
\centering
\begin{subfigure}{.3\textwidth}
  \centering
  \includegraphics[width=\textwidth]{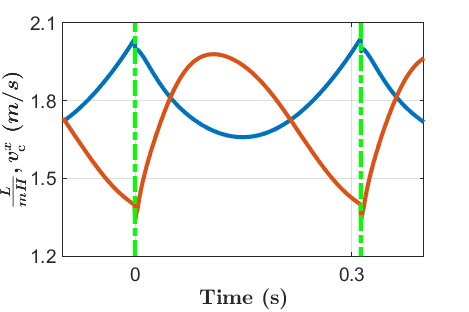}
  \caption{Rabbit \textcolor{blue}{$ \frac{ L}{mH}$} and \textcolor{red}{$v_{\rm c}^x$}}
\end{subfigure}
\begin{subfigure}{.3\textwidth}
  \centering
  \includegraphics[width=\textwidth]{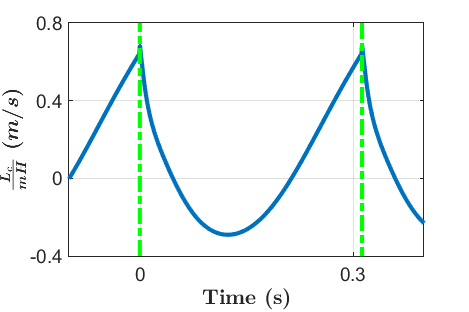}
  \caption{Rabbit $\frac{L_{\rm c}}{mH}$}
\end{subfigure}
\begin{subfigure}{.3\textwidth}
  \centering
  \includegraphics[width=\textwidth]{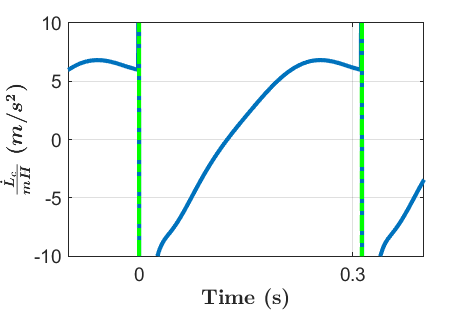}
    \caption{Rabbit $\frac{\dot{L}_{\rm c}}{mH}$}
\end{subfigure}

\begin{subfigure}{.3\textwidth}
  \centering
  \includegraphics[width=\textwidth]{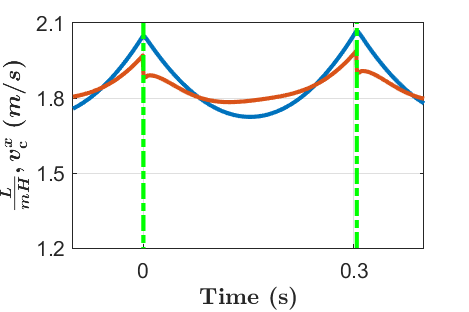}
  \caption{Cassie \textcolor{blue}{$ \frac{ L}{mH}$} and \textcolor{red}{$v_{\rm c}$}}
\end{subfigure}
\begin{subfigure}{.3\textwidth}
  \centering
  \includegraphics[width=\textwidth]{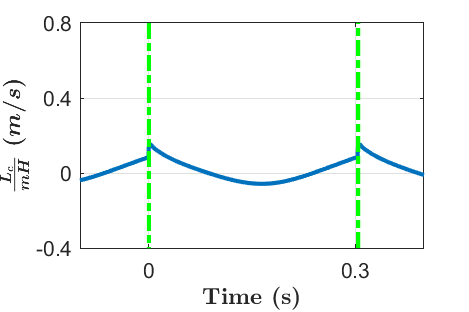}
  \caption{Cassie $\frac{L_{\rm c}}{mH}$}
\end{subfigure}
\begin{subfigure}{.3\textwidth}
  \centering
  \includegraphics[width=\textwidth]{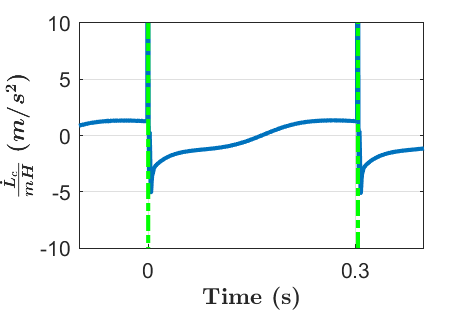}
  \caption{Cassie $\frac{\dot{L}_{\rm c}}{mH}$}
\end{subfigure}
\caption{Plots of $L$, $v_{\rm c}$, and $\dot{L}_{\rm c}$ for the bipedal robots Rabbit and Cassie walking at about 2m/s, while $z_{\rm c}$ is carefully regulated to 0.6m. The vertical green lines indicate the moment of impact. For both robots, the angular momentum about the contact point, $L$, has a convex shape \chg{(due to $\dot{L} = mgx_{\rm c} + u_a$, $u_a = 0$ and CoM passes the contact point only once)}, similar to the trajectory of a LIP model, while the trajectory of the longitudinal velocity of the center of mass, $v_{\rm c}$, has no consistent shape. The variation of $L_{\rm c}$ throughout a step, which is caused by the legs of the robot having mass, is what leads to a difference in the CoM velocity between a real robot and a LIP model. \chg{The patterns of $L_{\rm c}$ shown above are not specific to certain robot or controller but match the walking mechanism described in \cite{MALIGRRUSE10,wieber2006holonomy}.} In this figure, $L$ is continuous at impact, which is based on two conditions: $v^z_{\rm c}=0$ at impact and the ground is level. Even when these two conditions are not met, the jump in $L$ at impact can be easily calculated with \eqref{AM_transfer}.}
\label{fig:comparison_of_L_V}
\end{figure*}
\begin{figure*}[ht]
\centering
\begin{subfigure}{\textwidth}
\label{fig:L_vc_Lc_dLc_FLW_a}
  \centering
  \includegraphics[width=1\textwidth]{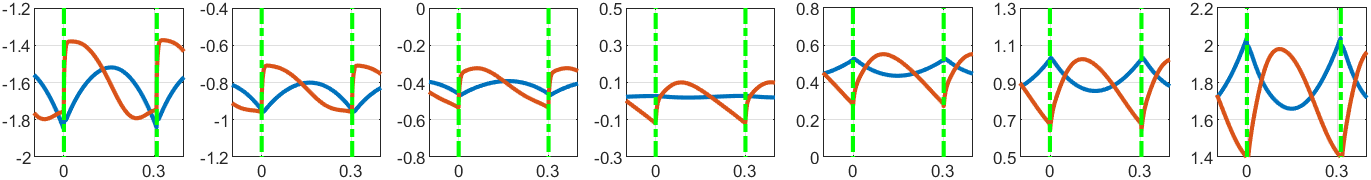}
  \caption{\textcolor{blue}{$ \frac{\mathbf L}{\mathbf{mH}}$} and \textcolor{red}{$\mathbf v_{\mathbf c}$} in $m/s$ versus time in seconds.}
\end{subfigure}
\begin{subfigure}{\textwidth}
\label{fig:L_vc_Lc_dLc_FLW_b}
  \centering
  \includegraphics[width=1\textwidth]{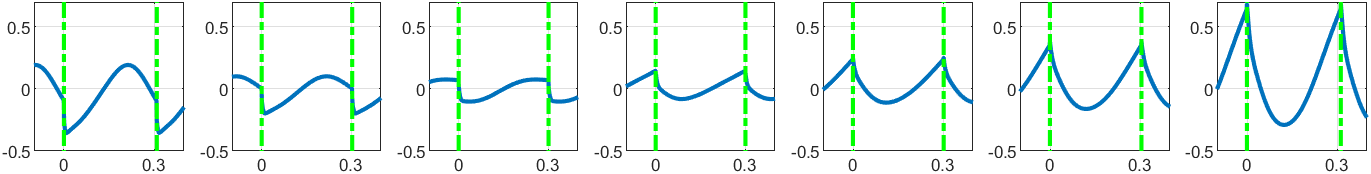}
  \caption{$\frac{{\mathbf L}_{\mathbf c}}{\mathbf{m H}}$ in $m/s$  versus time in seconds.}
\end{subfigure}
\begin{subfigure}{\textwidth}
\label{fig:L_vc_Lc_dLc_FLW_c}
  \centering
  \includegraphics[width=1\textwidth]{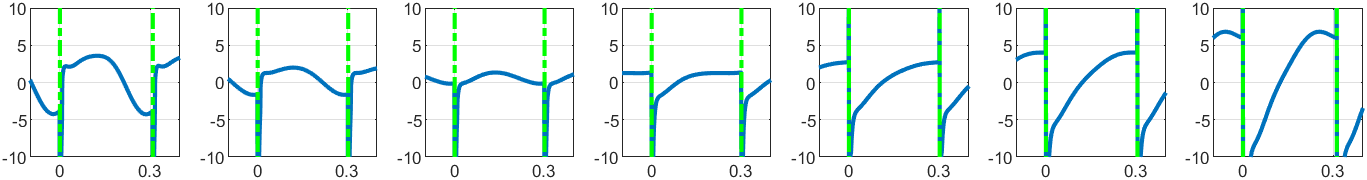}
  \caption{ $\frac{ \stackrel{\bullet}{\mathbf L}_{\mathbf c}}{\mathbf{mH}}$ in $m/s^2$  versus time in seconds. }
\end{subfigure}
\caption{Plots of $L$, $v_{\rm c}$, and $\dot{L}{\rm c}$ for Rabbit walking at different speeds. The green vertical lines indicate the moment of impact. (a) shows that $L$ always has a convex or concave shape like the LIP model, while $v_{\rm c}$ has no determinant shape. The shape of $L$ is a direct consequence of $\dot{L} = mgx{\rm c}$.
The quantities \textcolor{blue}{$ \frac{\mathbf L}{\mathbf{mH}}$} and \textcolor{red}{$\mathbf v_{\mathbf c}$} are close in scale and oscillate about one another. This shows that  directly regulating $L$ does indeed indirectly regulate $v_{\rm c}$. (b) and (c) show the scales of $L_{\rm c}$ and $\dot{L}{\rm c}$. It is seen that $\dot{L}{\rm c}$ is much larger in scale and thus omitting it in \eqref{eqn:zdOttConstantHeight} can create a larger error than neglecting $L_{\rm c}$ in \eqref{eqn:ZDnaturalCoordinatesConstantHeight}.}
\label{fig:L_vc_Lc_dLc_FLW}
\end{figure*}

What are there advantages to using $L$?
\begin{enumerate}
\renewcommand{\labelenumi}{(\alph{enumi})}
\setlength{\itemsep}{.2cm}
    \item The first advantage of controlling $L$ is that it provides a more comprehensive representation of current walking status because it is the sum of angular momentum about the center of mass, $L_{\rm c}$,  and linear  momentum, $p_{\rm c}\wedge m v_{\rm c}$. From \eqref{eqn:zdOtt}, we see that there exists momentum transfer between these two quantities. If $L_{\rm c}$ increases, it must ``take'' some momentum away from $v_{\rm c}$, and vice versa. \ygout{By the nature of walking, unless a flywheel is installed on the robot, $L_{\rm c}$ must oscillate about zero during a normal step for otherwise, some joints will likely hit their joint limit.} \chg{For normal bipedal walking, $L_{\rm c}$ oscillates about zero.} $L_{\rm c}$ functions to store momentum\cite{pratt2006velocity}, but importantly it can an only store it for a short amount of time. When designing a foot placement strategy, it is important to take the ``stored'' momentum into account. 
    
    When balancing on one foot for example, some researchers plan $L_{\rm c}$ and $v_{\rm c}$ separately \cite{dai2016planning}, or use $L_{\rm c}$ as an input to regulate balance by waving the torso, arms, or swing leg\cite{hofmann2009exploiting,goswami2004rate} or even a flywheel \cite{xiong2020sequential}. Here, instead of moving limbs to generate a certain value of $L_{\rm c}$, we view $L_{\rm c}$ as a result of the legs and torso moving to fulfill other tasks.  \chg{In this paper, we observe $L_{\rm c}$ and take it into consideration through $L$ and do not seek to regulate it directly as an independent quantity.}  
    
    \begin{figure}
        \centering
        \includegraphics[width = 0.45\textwidth]{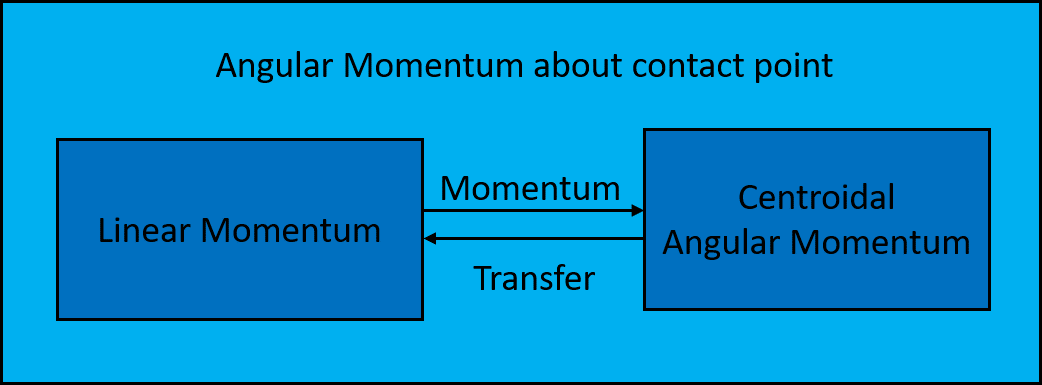}
        \caption{The relation between $L$, $L_{\rm c}$, and $v^x_{\rm c}$. Equation~\eqref{eq:AngularMomentumLinearMomentum} shows L is the sum of $L_{\rm c}$ and a term that is linear in $v_{\rm c}$, while the second line of \eqref{eqn:zdOtt} shows the transfer of momentum between $L_{\rm c}$ and $v_{\rm c}$. The relation is an analogue of mechanical, kinetic and potential energy.}
        \label{fig:L_Lc_vc_relation}
    \end{figure}

    \item  Secondly, because $\dot{L}= mgx_{\rm c} + u_a $  depends only on the CoM position, it follows that $L$ has relative degree three \chg{with respect to all inputs except the stance ankle torque, where it has relative degree one}. Consequently, the evolution $L$ is only weakly affected by motor torques of the body, that is $u_b$, during a step. In Fig. 
    \ref{fig:comparison_of_L_V} (a) and (d) and Fig.~\ref{fig:L_vc_Lc_dLc_FLW} (a) we see that the trajectory of $L$ consistently has a convex shape when stance ankle torque is zero, irrespective of model or speed. \chg{We'll see later the same property in experimental data.}
    
    \item 
    The discussion so far has focused on the single support phase of a walking gait. Bipedal walking is characterized by the transition between left and right legs as they alternately take on the role of stance leg (aka support leg) and swing leg (aka non-stance leg). In double support, the transfer of angular momentum between the two contact points satisfies
    \begin{equation}\label{AM_transfer}
        L_{2} =  L_{1} + p_{2\to 1}\wedge m v_{c},
    \end{equation}
    where $p_{2\to 1}$ is the vector from point 2 (the new stance leg position) to point 1 (the previous stance leg position). 
    Hence, the change of angular momentum between two contact points depends only on the vector defined by the two contact points and the center of mass velocity. In particular, angular momentum about a given contact point is invariant under the impulsive force generated at that  contact point. Consequently, we can easily determine the angular momentum about the new contact point by \eqref{AM_transfer} when impact happens without resorting to approximating assumptions about the impact model. Moreover, if $\dot{z}{\rm c}$ is zero and the ground is level, then $p_{2\to 1} \wedge m v_{c}=0$, and hence $L_2 = L_1$. \chg{We note that pendulum models parameterized with CoM velocity often assume continuity at impact, which is not generally true for real robots.}
\end{enumerate}  

Figure \ref{fig:comparison_of_L_V} shows the evolution of $L$, $v_{\rm c}$, $L_{\rm c}$ and $\dot{L}_{\rm c}$ during a step for both Cassie and Rabbit, when walking speed is about 2 m/s, $\dot{z}_{\rm c}=0$, and no stance ankle torque is applied. Figure \ref{fig:L_vc_Lc_dLc_FLW} shows the evolution of $L$, $v_{\rm c}$, $L_{\rm c}$ for Rabbit walking at a range of speeds from -1.8 m/s to 2.0 m/s. Figure~\ref{fig:comparison_of_L_V} and \ref{fig:L_vc_Lc_dLc_FLW} also show the continuity property of $L$ at impact.

\section{Comparison of Approximate Models for Center of Mass Dynamics}
\label{sec:PendulumModels}
Each of the \ygout{representations of the essential dynamics in} \chg{dynamical models} \eqref{eqn:zdOtt}, \eqref{eqn:ZDnaturalCoordinates}, and \eqref{eqn:JWG:ZD_theta_c} is valid along all trajectories of the full-dimensional model. This section systematically goes through the models in Sec.~\ref{sec:LowDimesionDynamics} and looks for connections with low-dimensional pendulum models.  Subsequently, Sect.~\ref{sec:ControlZDoffzeroDynamics} makes connections between pendulum models and the zero dynamics. 

\subsection{Constant Pendulum Height}
\ygout{We suppose one component of the virtual constraints in \eqref{eq:VCeqn} is $z_{\rm c}(q) - H $, where $H$ is a constant. Then $y\equiv 0$  yields ${z}_{\rm c}=H$, $\dot{z}_{\rm c}=0$, and $\ddot{z}_{\rm c}=0$, simplifying \eqref{eqn:zdOtt} and \eqref{eqn:ZDnaturalCoordinates} to}

\chg{If CoM height is constant, i.e., ${z}_{\rm c}=H$, $\dot{z}_{\rm c}=0$, and $\ddot{z}_{\rm c}=0$, then \eqref{eqn:zdOtt} and \eqref{eqn:ZDnaturalCoordinates} become}
\chg{
\begin{equation}
\label{eqn:zdOttConstantHeight}
    \begin{aligned}
      \dot{x}_{\rm c}&= v_{\rm c}\\
      \dot{v}^x_{\rm c}&=\frac{g}{H} x_{\rm c}  -\frac{\dot{L}_{\rm c}}{m H} + \frac{u_a}{mH},
    \end{aligned}
\end{equation}
and
\begin{equation}
    \label{eqn:ZDnaturalCoordinatesConstantHeight}
    \begin{aligned}
       \dot{x}_{\rm c}&= \frac{L}{m H} - \frac{L_{\rm c}}{m H}  \\
       \dot{L}&=m g x_{\rm c} + u_a,
    \end{aligned}
\end{equation}
respectively. Equation \eqref{eqn:ZDnaturalCoordinatesConstantHeight} can be rewritten as 
\begin{equation}
    \label{eqn:ZDnaturalCoordinatesConstantHeight_2}
    \begin{aligned}
       \dot{x}_{\rm c}&= v_p - \frac{L_{\rm c}}{m H} \\
       \dot{v}_p&=\frac{g}{H} x_{\rm c} + \frac{u_a}{mH},
    \end{aligned}
\end{equation}
}

\noindent where $v_p = \frac{L}{mH}$, which is more directly comparable to \eqref{eqn:zdOttConstantHeight}. In this paper we frequently plot $L$ scaled by the coefficient $\frac{1}{mH}$, so that it can be more directly compared to $v_{\rm c}$ (same units and similar magnitudes).

At this point, no approximations have been made and both models are valid everywhere that $z_{\rm c}(q) \equiv H$. Hence, the two models are still \textit{equivalent} representations of the center of mass dynamics for all trajectories  satisfying $z_{\rm c}(q) \equiv  H$. \textit{We'll next argue that the models are not equivalent when it comes to approximations.} 

Dropping the $\dot{L}_{\rm c}$ term in \eqref{eqn:zdOttConstantHeight} results in:
\chg{
\begin{equation}
    \label{eqn:JWG:ZD_LIP}
    \begin{aligned}
      \dot{x}_{\rm c}&= v_{\rm c} \\
        \dot{v}_{\rm c}&=\frac{g}{H} x_{\rm c} + \frac{u_a}{mH}.
    \end{aligned}
\end{equation}
}
This is the well-known LIP model proposed by \cite{KATA91}.

Dropping $L_{\rm c}$ in \eqref{eqn:ZDnaturalCoordinatesConstantHeight} results in 
\begin{equation}
    \label{eqn:JWG:ZD_ALIP}
    \begin{aligned}
      \dot{x}_{\rm c}&= \frac{L}{m H} \\
      \dot{L}&=m g x_{\rm c} +  u_a,
    \end{aligned}
\end{equation}
\noindent which is used in \cite{powell2016mechanics,gong2021one}. \chg{In \cite{powell2016mechanics}, \eqref{eqn:JWG:ZD_ALIP} is used instead of \eqref{eqn:JWG:ZD_LIP} so that the easy ``update'' property of $L$ at impact can be used. In this paper, we demonstrate that during the continuous phase, the states of \eqref{eqn:JWG:ZD_ALIP} much more accurately capture the evolution of $(x_{\rm c}, L)$ in a real robot than the states of \eqref{eqn:JWG:ZD_LIP} capture the evolution of $(x_{\rm c}, v_{\rm c})$. Moreover, we will make use of this improved accuracy in the design of a feedback controller}.  To distinguish the model \eqref{eqn:JWG:ZD_ALIP} from \eqref{eqn:JWG:ZD_LIP}, we will denote it by ALIP, where A stands for Angular Momentum.

For a robot with a point mass, the two models \eqref{eqn:JWG:ZD_LIP} and \eqref{eqn:JWG:ZD_ALIP} are equivalent\chg{, because $L_{\rm c}$ is then identically zero}. For a real robot with $L_{\rm c}$ and $\dot{L}_{\rm c}$ that are nonnegligible, however, we argue that \eqref{eqn:JWG:ZD_ALIP} is more accurate than \eqref{eqn:JWG:ZD_LIP} primarily because of three properties,
\begin{enumerate}
\renewcommand{\labelenumi}{(\alph{enumi})}
\setlength{\itemsep}{.2cm}
\item \chg{Relative Amplitude. Based on our observations, the ratio of $\dot{L}_{\rm c}/mgx_{\rm c}$ is much larger than $L_{\rm c}/L$ over a wide range of walking velocities; thus the simplification \eqref{eqn:JWG:ZD_ALIP} introduces relatively less error than \eqref{eqn:JWG:ZD_LIP}}.
\item Relative degree. $L$ has relative degree two with respect to $L_{\rm c}$ and three with respect to $\dot{L}_{\rm c}$, whereas $v_{\rm c}$ has relative degree one with respect to $\dot{L}_{\rm c}$.  \chg{Because integration is a form of low-pass filtering,} the lower relative degree makes $v_{\rm c}$ more sensitive to the omission of the $L_{\rm c}$ term.
\item $L_{\rm c}$ oscillates about zero. What makes \eqref{eqn:JWG:ZD_ALIP} even more accurate is that, \chg{based on our own observation and references \cite{MALIGRRUSE10,wieber2006holonomy}, the sagittal plane component of $L_{\rm c}$ oscillates about zero for periodic and non-periodic gaits.} \ygout{and, moreover, its average value over a step is close to zero as shown in (Table removed). Though the data we show here corresponds to the specific controller described in Sec \ref{Sec:Controller} and \ref{Sec:Experiment}, we believe it is generally true for other controllers that induce periodic walking.} The oscillation of $L_{\rm c}$  results in the effect of $L_{\rm c}$ on $x_{\rm c}$ roughly averaging out to zero over a step. \ygout{If we seek to predict the value of $L$ with \eqref{eqn:JWG:ZD_ALIP} for a longer period of time, the effect of $L_{\rm c}$ will cancel out because of the oscillation. If we seek to predict $L$ for a shorter time interval, the error will be small because the time duration is short.}
\end{enumerate}

\begin{figure*}
\centering
\begin{subfigure}{0.245\textwidth}
  \centering
  \includegraphics[width=\textwidth]{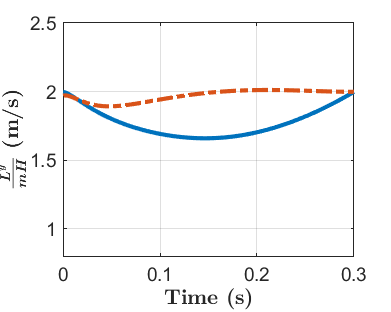}
  \caption{Rabbit $L^y$ prediction}
  \label{fig:L_prediction_Rabbit}
\end{subfigure}
\begin{subfigure}{0.245\textwidth}
  \centering
  \includegraphics[width=\textwidth]{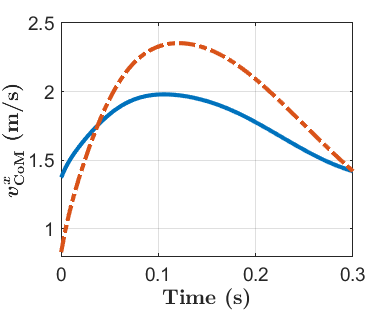}
  \caption{Rabbit $v_{\rm c}$ prediction}
  \label{fig:vx_prediction_Rabbit}
\end{subfigure}
\begin{subfigure}{0.245\textwidth}
  \centering
  \includegraphics[width=\textwidth]{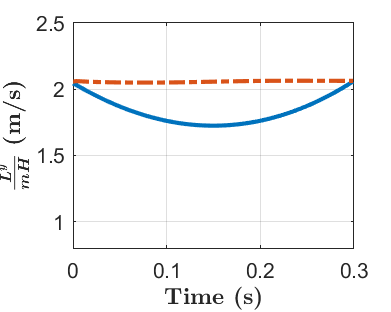}
  \caption{Cassie $L^y$ prediction}
  \label{fig:L_prediction_Cassie}
\end{subfigure}
\begin{subfigure}{0.245\textwidth}
  \centering
  \includegraphics[width=\textwidth]{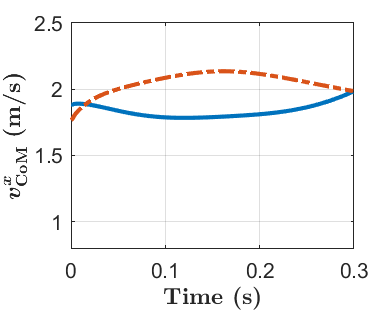}
  \caption{Cassie $v_{\rm c}$ prediction}
  \label{fig:vx_prediction_Cassie}
\end{subfigure}
\caption{Comparison of the ability to predict velocity vs angular momentum at the end of a step. The instantaneous values are shown in \textcolor{blue}{\bf blue} and the predicted value at the end of the step is shown in \textcolor{red}{\bf red}, where a perfect prediction would be a flat line that intercepts the terminal point of the blue line. The most crucial decision in the control of a bipedal robot is where to place the next foot fall. In the standard LIP controller, the decision is based on predicting the longitudinal velocity of the center of mass. In Sect.~\ref{Sec:Controller} we use angular momentum about the contact point. We do this because on realistic bipeds, a LIP-style model provides a more accurate and reliable prediction of $L$ than $v_{\rm c}$. The comparison is more significant on Rabbit, whose leg center of mass is further away from the overall center of mass. }
\label{fig:LIP_prediction}
\end{figure*}

In Fig.~\ref{fig:LIP_prediction}, we have used the models \eqref{eqn:JWG:ZD_LIP} and \eqref{eqn:JWG:ZD_ALIP} to predict the values of $v_{\rm c}$ and $L$ at the end of a step. We plot $\frac{L}{mH}$ instead of $L$ to make the scale and units comparable. The blue line is the true trajectory of $L$ (resp.~$v_{\rm c}$) during a step. The red line shows the prediction of $L$ (resp.~$v_{\rm c}$) at the end of each of step, at each moment throughout a step, based on the instantaneous values of $x_{\rm c}$ and $L$ ($v_{\rm c}$) at that moment. The red line would be perfectly flat if \eqref{eqn:JWG:ZD_ALIP} and \eqref{eqn:JWG:ZD_LIP} perfectly captured the evolution of $L$ ($v_{\rm c}$), respectively, in the full simulation model, and the flatter the estimate, the more faithful is the representation. 

\begin{figure}
    \centering
    \includegraphics[width=0.4\textwidth]{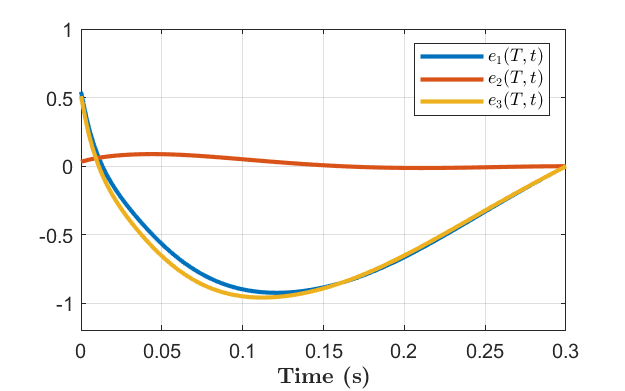}
    \caption{A plot of the error terms in \eqref{eq:Estimation_Error_1} and \eqref{eq:Estimation_Error_2} resulting from dropping $\dot{L}_{\rm c}$ and $L_{\rm c}$, respectively, for the Rabbit model walking at 2 m/s. The take-home message is that of the terms $e_2(t_1, t_2)+e_3(t_1, t_2)$ in \eqref{eq:Estimation_Error_1} comprising the velocity error of the LIP model, the term $e_3(t_1, t_2)$ shown in the \textcolor{Goldenrod}{\bf yellow} line contributes by far the largest portion of the total error shown by the \textcolor{blue}{\bf blue} line. The error of the ALIP model, however, is given only by $e_2(t_1, t_2)$, which results in the significantly reduced prediction error shown by the \textcolor{red}{\bf red} line.}
    \label{fig:ErrorTerms}
    \vspace*{-.5cm}
\end{figure}

\begin{figure}[ht]
\centering
\begin{subfigure}{.35\textwidth}
  \centering
  \includegraphics[width=\textwidth]{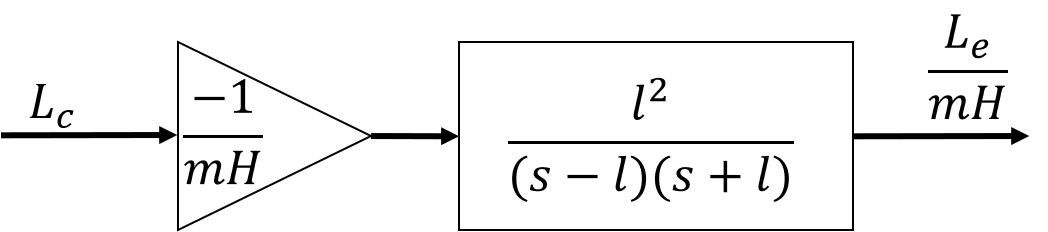}
  \caption{ALIPM}
  \vspace*{.2cm}
\end{subfigure}
\begin{subfigure}{.35\textwidth}
  \centering
  \includegraphics[width=\textwidth]{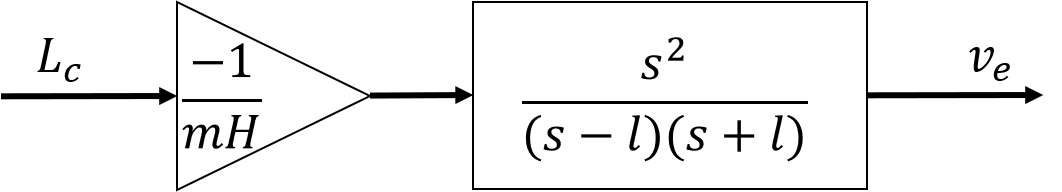}
  \caption{LIPM}
\end{subfigure}
\begin{subfigure}{.45\textwidth}
  \centering
  \includegraphics[width=\textwidth]{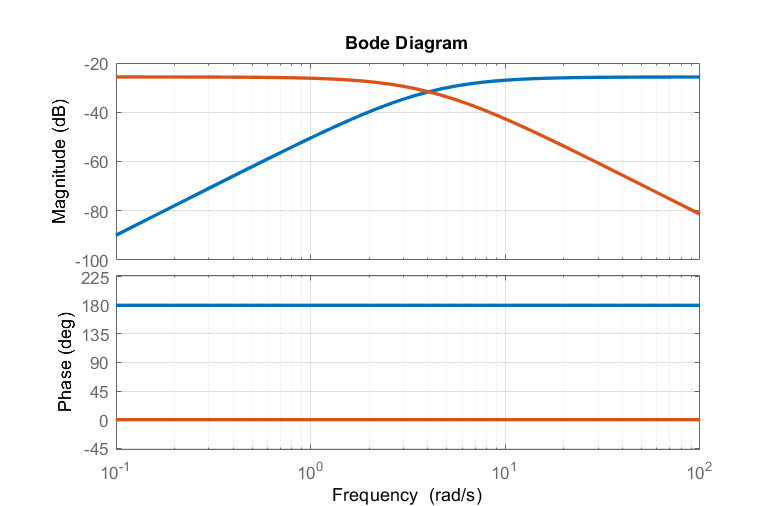}
  \caption{Bode Plot}
\end{subfigure}
\caption{How neglecting $L_{\rm c}$ and $\dot{L}_{\rm c}$ generates errors in ALIPM and LIPM. Note the low-pass (ALIPM in  \textcolor{red}{\bf red}) vs high-pass (LIPM in  \textcolor{blue}{\bf blue}) nature of the respective transfer functions.  }
\label{fig:Laplace_Tranform_Bode}
\end{figure}

The prediction errors of \eqref{eqn:JWG:ZD_LIP} and \eqref{eqn:JWG:ZD_ALIP} caused by neglecting $L_{\rm c}$ and $\dot{L}_{\rm c}$, respectively, satisfy
\chg{
\begin{equation}
\label{eqn:ErrorVel}
    \begin{aligned}
      \dot{x}_e&= v_e\\
      \dot{v}_e&=\frac{g}{H} x_e  -\frac{\dot{L}_{\rm c}}{m H},
    \end{aligned}
\end{equation}
and
\begin{equation}
    \label{eqn:ErrorAM}
    \begin{aligned}
       \dot{x}_e &= \frac{L_e}{m H} - \frac{L_{\rm c}}{m H} \\
       \dot{L}_e&=m g x_e,
    \end{aligned}
\end{equation}
 where $(x_e, L_e)$ are the differences in the trajectories of  \eqref{eqn:JWG:ZD_LIP} and \eqref{eqn:ZDnaturalCoordinatesConstantHeight_2}; similarly, $(x_e, v_e)$ are the differences in the trajectories of  \eqref{eqn:JWG:ZD_ALIP} and \eqref{eqn:zdOttConstantHeight}. Direct solution of these two sets of differential equations for zero initial conditions leads to }
\begin{align}
    \label{eq:Estimation_Error_1}
    v_e(t_2,t_1) &=  e_1(t_2,t_1) \nonumber \\
    &=  e_2(t_2,t_1) + e_3(t_2,t_1)  \\ 
    \label{eq:Estimation_Error_2}
   \frac{L_e(t_2,t_1)}{mH} &=  e_2(t_2,t_1),
\end{align}
where 
\begin{equation}\label{eq:ErrorTerms}
\begin{aligned}
    e_1(t_2,t_1) =& - \frac{1}{mH}\int_{t_1}^{t_2}\cosh(\ell(t_2-\tau))\dot{L}_{\rm c}(\tau)\,d\tau \\
    e_2(t_2,t_1) =& - \frac{1}{mH}\int_{t_1}^{t_2} \ell\sinh(\ell(t_2-\tau))L_{c}(\tau)\,d\tau \\
    e_3(t_2,t_1) =&- \frac{1}{mH}\big(L_{\rm c}(t_2) - \cosh(\ell (t_2-t_1))L_{c}(t_1) \big). \nonumber \\
\end{aligned}
\end{equation}
Figure~\ref{fig:ErrorTerms} shows the (relative) sizes of these error terms. 
If we view $L_{\rm c}$ as a disturbance and prediction error as an output in \eqref{eqn:ErrorVel} and \eqref{eqn:ErrorAM}, we obtain the corresponding Laplace transforms and Bode plots shown in Fig.~\ref{fig:Laplace_Tranform_Bode}.


\begin{figure}[ht]
\centering
\begin{subfigure}{.45\textwidth}
  \centering
  \includegraphics[width=\textwidth]{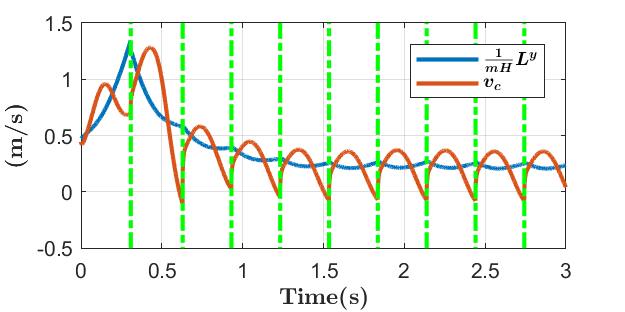}
  \caption{LIP controller}
  \vspace*{.2cm}
\end{subfigure}
\begin{subfigure}{.45\textwidth}
  \centering
  \includegraphics[width=\textwidth]{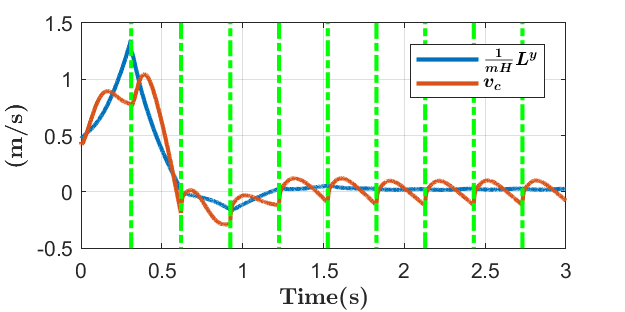}
  \caption{ALIP controller}
\end{subfigure}
\caption{\chg{Simulation results of Rabbit with controllers based on LIP and ALIP, following an identical design philosophy, based on foot placement. The details of the controller are described in Sec \ref{Sec:Controller} and Sec \ref{sec:virtual_constraint_Cassie}. The controller based on the ALIP model is much more effective in regulating velocity to zero.}}
\label{fig:sim_compare}
\end{figure}

\chg{
\subsection{Simulation Comparison}
We compare controllers designed on the basis of the ALIP and LIP models in simulation. The results shown in Fig \ref{fig:sim_compare} demonstrate the advantage of using ALIP over LIP for controller design. The initial hip velocity is set to 0.5 m/s and hip position is centered over the contact point. The goal of each controller is to regulate $v_{\rm c}$ (resp., $L$) to zero, with foot placement as the decision variable and step duration constant. In the plots, we observe that the ALIP-based controller regulates $L$ closely to zero and thus has an average $v_{\rm c}$ close to zero, while the LIP-based controller is unable to regulate $v_{\rm c}$ effectively. The reason is that, at the end of a step, the linear momentum was transferred to centroidal angular momentum $L_{\rm c}$ due to the movement of Rabbit's heavy legs (see Eqn \eqref{eqn:ZDnaturalCoordinatesConstantHeight_2}), resulting in a small $v_{\rm c}$, which misleads the LIP controller into choosing a small foot displacement. In the ALIP model, $L$ is less affected by momentum transfer between $v_{\rm c}$ and $L_{\rm c}$ because $L$ captures their sum, and thus the ALIP model suggests better foot placement. Though with a LIP controller it is possible to regulate velocity through ZMP (ankle torque) during continuous phase, we argue that with an ALIP controller, the capability of ZMP can be reserved for better purposes than compensating for model error.
}

\ygout{Accounting for $L_{\rm c}$}

\begin{figure}[ht]
\centering
\begin{subfigure}{.24\textwidth}
  \centering
  \includegraphics[width=\textwidth]{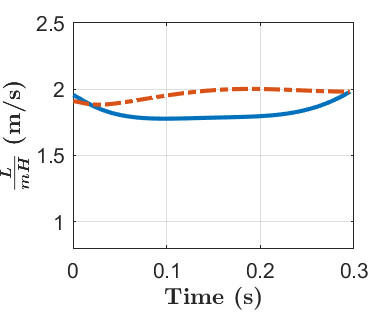}
  \caption{Rabbit $L$ prediction}
\end{subfigure}
\begin{subfigure}{.24\textwidth}
  \centering
  \includegraphics[width=\textwidth]{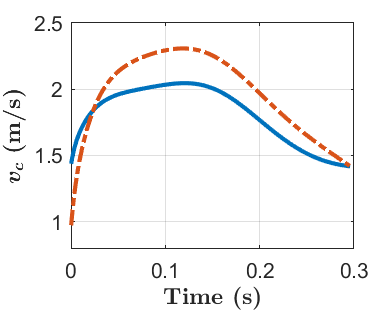}
  \caption{Rabbit $v_{\rm c}$ prediction}
\end{subfigure}
    \caption{\chg{Comparison of the ability to predict velocity vs angular momentum at the end of a step in a model with ankle torque $u_a = 30\sin(2\pi \tau/T)$, where $\tau$ varies from 0 to $T$ during a step. The instantaneous values are shown in \textcolor{blue}{\bf blue} and the predicted value at the end of the step is shown in \textcolor{red}{\bf red}. Because ankle torque is an input, we assume its trajectory is known when making predictions. For comparison purposes with Fig. \ref{fig:LIP_prediction}, the ankle torque is chosen to be sufficiently large so that gravity is no longer dominant in $\dot{L} = mgx_{\rm c} + u_a$ and the trajectory of $L$ is no longer convex.}}
\label{fig:pred_ua}
\end{figure}

\chgcomment{whole subsection removed}
\chg{
\subsection{ Non-zero Ankle Torque}
In previous subsections we have demonstrated the accuracy of pendulum model parameterized with $L$ when ankle torque is zero. According to Eqn \eqref{eqn:ZDnaturalCoordinates}, the effect of $L_{\rm c}$ and $u_a$ on the system are independent due to the superposition property. So if dropping $L_{\rm c}$ term has little effect on the model accuracy when $u_a$ is zero, it should still has little effect on the model accuracy when $u_a$ is non-zero. Though $L_{\rm c}$ trajectory itself will be changed when $u_a$ is non-zero, its pattern is still similar. Here for completeness we run a simulation on Rabbit. The results are shown in Fig. \ref{fig:pred_ua}} 

\section{Stabilizing the ALIP Model} \label{Sec:Controller}

\ygout{ In this section, we explain our method for deciding where to end one step by initiating contact between the ground and the swing foot, thereby beginning the next step. In robot locomotion, this is typically called \mbox{\footnote{What else is there to control, one might ask? The overall posture of the robot is very important too.}} ``foot placement control''.Because of the advantages of angular momentum versus linear velocity that we listed in Sect.~\mbox{\ref{sec:AngularMomentumGeneral}} and \mbox{\ref{sec:PendulumModels}}, we will use angular momentum about the contact point as the primary control variable. To make things as clear as possible, we'll base the controller on the ALIP model \mbox{\eqref{eqn:JWG:ZD_ALIP}} which is a linear and time-invariant approximation of the zero dynamics. In Appendix~\mbox{(section removed)}, we provide a similar result for nonlinear and time-varying zero dynamics.}
 

 \chg{In this section, we provide a means to regulate angular momentum about the contact point to approximately achieve a desired walking speed. Specifically, the ALIP model \eqref{eqn:JWG:ZD_ALIP} is used to form a one-step ahead prediction of angular momentum, $L$. In combination with the angular momentum transfer formula \eqref{AM_transfer}, a feedback law results for where to place the swing foot at the end of the current step so as to achieve a desired angular momentum at the end of the ensuing step. In robot locomotion, this is typically called ``foot placement control'' \cite{RAI84,PRCADRGO06,da20162d,xiong2019orbit}. }
 
 \chg{$L$ can also can be controlled through vertical CoM velocity at the end of a step \cite{powell2016mechanics}, adjusting the step duration \cite{KATA91,khadiv2020walking}, ankle torque during continuous phase \cite{SAFU90}, which has a relative degree one relation to $L$, or by other torques applied at the body coordinates during the continuous single support phase \cite{grizzle2005nonlinear,azad2013angular,gonzalez2020line}, which have a relative degree three relation to $L$. We summarize these general methods\footnote{\chg{With the exception of the method based on regulating $L_c$, each of the methods listed in Table~\ref{tab:HowRegulateL} can be applied to a pendulum model parameterized with $L$. For humans, $L_c$ is mainly utilized when we are about to fall off a support structure, such as when balancing on a tightrope. In bipedal robots without a flywheel, the effect of $L_c$ on $L$ is weak, due to multiple factors such as kinematic constraints, torque limits, and being relative degree three with respect to body torques. $L_c$ is an ineffective means of regulating $L$ for the same reasons that dropping $L_c$ in the ALIP model does not introduce significant inaccuracy.}} to regulate $L$ in Table~\ref{tab:HowRegulateL}.}

 \begin{table}[h!]
 
 \center
 \begin{tabular}{ |l|l|}
 \hline
 \multicolumn{2}{|l|}{\textbf{Continuous Phase}}\\
 \hline
 Relative Degree Three   & $L_c$, $z_c$\\
 \hline
 Relative Degree One   & $u_a$\\
 \hline
 \multicolumn{2}{|l|}{\textbf{Transition Event}}\\
 \hline
 Initial condition for next continuous phase   & $x_c(0)$, $L(0)$\\
 \hline
 Phase duration & Step Time\\
 \hline
\end{tabular}
\caption{Methods to regulate $L$ for walking.}
\label{tab:HowRegulateL}
 \end{table}

 \subsection{Gait assumptions} 
\ygout{We make the following additional assumptions on the virtual constraints in \eqref{eq:VCeqn}, beyond those already made in \eqref{eq:VCeqn03} and \eqref{eq:VCeqn04}: }
\chg{When designing the foot placement controller, we assume the gait of the robot is controlled such that:}
\begin{enumerate}
\renewcommand{\labelenumi}{(\alph{enumi})}
    \item  the height of the center of mass is constant, that is $z_c \equiv H>0$;
    \item each step has constant duration $T>0$; and
    \item a desired swing leg horizontal position, $p^{x\, {\rm des}}_{\rm sw \to CoM}$, can be achieved at the end of the step.
\end{enumerate} 
We'll explain how to accomplish these objectives via the method of virtual constraints in Sections~\ref{sec:ControlZDoffzeroDynamics} and \ref{sec:virtual_constraint_Cassie}. 

\subsection{Notation}

\begin{figure}
    \centering
    \includegraphics[width=0.5\textwidth]{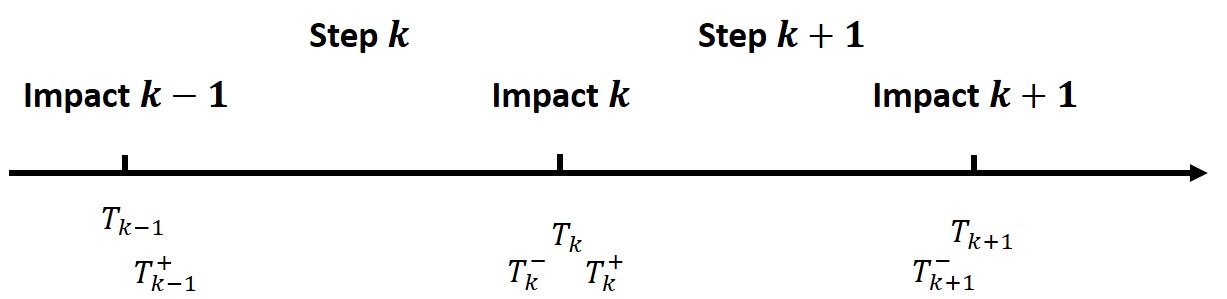}
    \caption{For a given time, $T_k$, the notation $T_k^-$ means that we are evaluating a function as a limit from the left of $T_k$, while $T_k^+$ means we are taking a limit from the right. This is compatible with how trajectories are defined for the hybrid model \eqref{eqn:grizzle:modeling:full_hybrid_model_walking}.}
    \label{fig:Def_Time}
\end{figure}

We distinguish among the following time instances when specifying the control variables.
\begin{itemize}
    \item $T$ is the step time.
    \item $T_k$ is the time of the $k$th impact and thus equals $k T$. 
    \item $T_k^-$ is the end time of step $k$, so that
    \item $T_k^+$ is the beginning time of step $k+1$ and $T_{k+1}^-$ is the end time of step $k+1$.
    \item $(T_k^- -t)=(T-\tau(t))$ is the time until the end of step $k$.
\end{itemize}
The superscripts $+$ and $-$ on $T_k$ are necessary because of the (potential) jump in a trajectory's values from the impact map; see \cite{westervelt2007feedback}. As shown in Fig.~\ref{fig:Def_Time}, $x(T_K^-)$ is the limit from the left of the model's solution at the time of impact, in other words it's value ``just before'' impact, while $x(T_K^+)$ is the limit from the right of the model's solution at the time of impact, in other words it's value ``just after'' impact.

\chg{With this notation, the reset map for the ALIP becomes
\begin{equation}
\label{eq:ALIPresetMAP}
\begin{aligned}
    x_c(T_k^+) &= p^{x}_{\rm sw \to CoM}(T_k^-) \\
    L(T_k^+)  &= L(T_k^-),
\end{aligned}
\end{equation}
after noting that \eqref{AM_transfer} simplifies to $L$ being constant across impacts when the ground is level and the vertical velocity of the center of mass is zero.}

We remind the reader that
\begin{itemize}
    \item $p_{\rm st \to CoM}$, $p_{\rm sw \to CoM}$ are the vectors emanating from stance/swing foot to the robot's center of mass. The stance foot defines the current contact point, while the swing foot is defining the point of contact for the next impact and is therefore a control variable.
    \item Also, for the implementation of the control law on the 3D biped Cassie in Sect.~\ref{sec:virtual_constraint_Cassie}, we need to distinguish between $L^y$ and $L^x$, the $y$ and $x$ components of the angular momentum (sagittal and frontal planes), respectively.
\end{itemize}

\subsection{Foot placement in longitudinal direction}
\label{sec:FootplacementLongitudinal}

 The control objective will be to place the swing foot at the end of the current step so as to achieve a desired value of angular momentum at the end of the ensuing step. The need to regulate the angular momentum one-step ahead of the current step, instead of during the current step, is because in \eqref{eqn:JWG:ZD_ALIP} $L$ is passive without ankle torque, in other words, it is not affected by the control actions of the current step. The only way to act on its states is through the transition events.

The closed-form solution of \eqref{eqn:JWG:ZD_ALIP} at time $T$ and initial time $t_0$ is
\begin{equation} \label{eq:LIP_AM_Preiction}
\footnotesize
\begin{bmatrix}
x_c(T) \\
L^y(T) \\
\end{bmatrix}
= A(T-t_0)
\begin{bmatrix}
x_c(t_0) \\
L^y(t_0)
\end{bmatrix},
\end{equation}
where 
$$A(t) = \begin{bmatrix}
\cosh(\ell t) & \sinh(\ell t)/(mH \ell) \\
mH\ell\sinh(\ell t) & \cosh(\ell t)
\end{bmatrix} $$
and  $\ell = \sqrt{\frac{g}{H}}$.

In the following, we breakdown the evolution of $L^y$ from $t$ to $T^-_{k+1}$, for three key time intervals or instances with the aim of forming a one-step-ahead estimate of angular momentum about the contact point.

\subsubsection{From \texorpdfstring{$t$}{TEXT} to \texorpdfstring{$T_k^-$}{TEXT}}

From the second row of \eqref{eq:LIP_AM_Preiction}, an \textit{estimate} for the angular momentum about the contact point at the end of current step, $\widehat{L}^y(T_k^-,t)$, can be continuously updated by

\begin{align} \label{eq:AM_this_end_predict}
    \widehat{L}^y(T_k^-,t) = & m H \ell \sinh(\ell(T_k^- - t))x_c(t)  \nonumber
    \\
    & + \cosh(\ell (T_k^- - t))L^y(t). 
\end{align}
Forming the running estimate in \eqref{eq:AM_this_end_predict}, versus a fixed estimate based on the values of $x_c$ and $L^y$ at the beginning of the step, allows disturbances to be taken into account.

\subsubsection{From \texorpdfstring{$T_k^-$}{TEXT} to \texorpdfstring{$T_k^+$}{TEXT}} 
This involves applying the reset map \eqref{eq:ALIPresetMAP}, yielding
\begin{align} 
\label{eq:rp_impact_1}
    x_c(T_k^+)&= p^x_{\rm sw \to CoM}(T_k^-) \\
\label{eq:AM_impact_relation_1}
    \widehat{L}^y(T_k^+,t)&= \widehat{L}^y(T_k^-,t).
\end{align}

\subsubsection{From \texorpdfstring{$T_k^+$}{TEXT} to \texorpdfstring{$T_{k+1}^-$}{TEXT}}
Similar to \eqref{eq:AM_this_end_predict}, the angular momentum at the end of the next step is estimated by
\begin{equation} \label{eq:AM_next_end_predict}
\small
\widehat{L}^y(T_{k+1}^-,t) = mH\ell\sinh(\ell T)x_c(T_k^+,t) + \cosh(\ell T)\widehat{L}^y(T_k^+,t). 
\end{equation}
Solving \eqref{eq:AM_this_end_predict}-\eqref{eq:AM_next_end_predict} so that 
$$\widehat{L}^y(T_{k+1}^-,t) = L^{y ~{\rm des}},$$
a desired value of angular momentum at the end of a step, \chg{(which can be obtained by $L^{y ~{\rm des}} = mHv^{x ~{\rm des}}$)}, yields a formula for the desired swing foot position at the end of the \textit{current} step, given the value of desired angular momentum at the end of the \textit{next} step,
\begin{equation} \label{eq:Desired_Footplacement}
    p^{x~{\rm des}}_{\rm \bf sw \to CoM}(T_k^-,t) :=\frac{L^{y ~{\rm des}} - \cosh(\ell T)\widehat{L}^y(T_k^-,t)}{mH \ell \sinh(\ell T)}.
\end{equation}

\chg{\noindent \textbf{Remark:} Instead of the deadbeat control \eqref{eq:Desired_Footplacement}, it is possible to asymptotically approach a desired value of $L^{\rm des}$ with the control law}
\chg{\begin{equation} \label{eq:Desired_FootplacementAsymptotic}
\begin{aligned}
    p^{x~{\rm des}}_{\rm \bf sw \to CoM}(T_k^-,t) &:=\frac{1 - \alpha }{mH \ell \sinh(\ell T)}L^{y ~{\rm des}} \\
   & +  \frac{\alpha - \cosh(\ell T)}{mH \ell \sinh(\ell T)} \widehat{L}^y(T_k^-,t),
    \end{aligned}
\end{equation}}

\chg{which achieves
\begin{equation}
( L^{y~{\rm des}}-\widehat{L}^y(T_{k+1}^-,t))   = \alpha (L^{y~{\rm des}}-\widehat{L}^y(T_{k}^-,t))
\end{equation}}
\chg{for $\alpha \in [0,1)$. Hence, for $\alpha = 0$, \eqref{eq:Desired_FootplacementAsymptotic} reduces to \eqref{eq:Desired_Footplacement}.}



\chg{\subsection{Stability Analysis of the ALIP for  \texorpdfstring{$L^{\rm des}$}{TEXT}}}

\chg{Consider the ALIP model \eqref{eqn:JWG:ZD_ALIP} with zero ankle torque and rest map \eqref{eq:ALIPresetMAP}. To compute the Poincar\'e map, we take the Poincar\'e section as $S:=\{(x_c, L, \tau)~|~ \tau=0^+\}$, which is the set of states just after impact. Computing \eqref{eqn:JWG:ZD_ALIP} over one step and using swing foot position with respect to the center of mass, $u_{fp}$, as an input, yields,
\begin{equation} \label{eq:discreteALIP}
\footnotesize
\begin{bmatrix}
x_c(T^+) \\
L(T^+) \\
\end{bmatrix}
= \begin{bmatrix}
0 & 0 \\
mH\ell\sinh(\ell T) & \cosh(\ell T)
\end{bmatrix}
\begin{bmatrix}
x_c(0^+) \\
L^(0^+)
\end{bmatrix}
+
\begin{bmatrix}
1 \\
0
\end{bmatrix}
u_{fp}(T^-).
\end{equation}
Next, applying the feedback law \eqref{eq:Desired_FootplacementAsymptotic} with $L^{\rm des}$ a constant results in the Poincar\'e map being
\begin{equation} \label{eq:PoincareALIP}
\footnotesize
\begin{aligned}
\begin{bmatrix}
x_c(T^+) \\
L(T^+) 
\end{bmatrix}
=& \begin{bmatrix}
\alpha-\cosh(\ell T) & \frac{(\alpha-\cosh(\ell T))\cosh(\ell T)}{mH\ell\sinh(\ell T)} \\
mH\ell\sinh(\ell T) & \cosh(\ell T)
\end{bmatrix}
\begin{bmatrix}
x_c(0^+) \\
L(0^+)
\end{bmatrix} \\
&+ \begin{bmatrix}
\frac{1-  \alpha}{mH\ell \sinh(\ell T)} \\
0
\end{bmatrix}
L^{des}.
\end{aligned}
\end{equation}
The Poincar\'e map has fixed point 
\begin{equation}
    \begin{bmatrix}
x_c^\ast\\
L^\ast 
\end{bmatrix} =     \begin{bmatrix}
\frac{1-\cosh(\ell T)}{mH\ell \sinh(\ell T)}L^{des} \\ L^{des}
\end{bmatrix}
\end{equation}
independent of $\alpha$ 
and the eigenvalues of the Poincar\'e map are $(\alpha, 0)$. Hence, for all $0 \le \alpha <1$, the fixed point is exponentially stable and moreover, \eqref{eq:PoincareALIP} is bounded-input bounded-state stable with respect to the command, $L^{\rm des}$.
}

\subsection{Lateral Control and Turning}
From \eqref{eqn:ZDnaturalCoordinatesConstantHeight}, the time evolution of the angular momentum about the contact point is decoupled about the $x$- and $y$-axes. Therefore, once a desired angular momentum at the end of next step is given, Lateral Control is essentially identical to Longitudinal Control and \eqref{eq:Desired_Footplacement} can be applied equally well in the lateral direction.\footnote{\chg{With a slight difference in the sign due to $L^x = -mgy_{\rm c}$.}} \textbf{The question becomes how to decide on  $\mathbf{L^{x~ \rm \bf des}(T_{k+1}^-)}$,} since it cannot be simply set to zero for walking with a non-zero stance width.

For walking in place or walking with zero average lateral velocity, it is sufficient to obtain $L^{x~ \rm des}$ from a periodically oscillating LIP model,
\begin{equation} \label{eq:lateral_desired_Lx}
    L^{x~ \rm des}(T_{k+1}^-) = \pm\frac{1}{2}mHW\frac{\ell\sinh(\ell T)}{1+\cosh(\ell T)},
\end{equation}
where $W$ is the desired step width. The sign is positive if next stance is left stance and negative if next stance is right stance. Lateral walking can be achieved by adding an offset to $L^{x~ \rm des}$. 

To enable turning, we assume a target direction is commanded and associate a frame to it by aligning the $x$-axis with the target direction while keeping the $z$-axis vertical. To achieve turning, we then define the desired angular momentum $L^{y~ \rm des}$ and $L^{x~ \rm des}$ in the new frame and use the hip yaw-motors to align the robot in that direction.

\section{Pendulum Models, Zero Dynamics, and Overall System Stability} 
\chgcomment{This whole section has been re-written. It is not possible to highlight subtractions and additions.}
\label{sec:ControlZDoffzeroDynamics}

This section establishes connections between the pendulum models of Sec.~\ref{sec:LowDimesionDynamics} and the swing phase zero dynamics as developed in \cite{WGCCM07}, or more precisely, approximations of the zero dynamics.
This is accomplished by analyzing how the zero dynamics are driven by the states of a bipedal robot's full-order model and its feedback controller when the closed-loop system is evolving off the zero dynamics manifold. As a main contribution, the analysis will yield conditions under which the driving terms are small and hence do not adversely affect the stability predictions associated with the exact zero dynamics. A secondary contribution of the section will be a presentation of the swing phase zero dynamics for a more general set of ``virtual constraints'' than those developed in \cite{WGCCM07,powell2013speed,griffin2015nonholonomic}.

\subsection{Intuitive Background}

An initial sense of the meaning and mathematical foundation of the swing phase zero dynamics can be gained by considering a floating-base model of a bipedal robot, and then its pinned model, that is, the model with a point or link of the robot, such as a leg end or foot, constrained to maintain a constant position respect to the ground. The given contact constraint is holonomic and constant rank, and thus using Lagrange multipliers (from the principle of virtual work), a reduced-order model compatible with the (holonomic) contact constraint is easily computed. When computing the reduced-order model, no approximations are involved, and solutions of the reduced-order model are solutions of the original floating-base model, with inputs (ground reaction forces and moments) determined by the Lagrange multiplier.

 Virtual constraints are relations (i.e., constraints) on the state variables of a robot's model that are achieved through the action of actuators and feedback control instead of physical contact forces. They are called \textit{virtual} because they can be re-programmed on the fly without modifying any physical connections among the links of the robot or its environment. We use virtual constraints to synchronize the evolution of a robot's links, so as to create exponentially stable motions. Like physical constraints, under certain regularity conditions, they induce an exact low-dimensional invariant model, called the \textit{zero dynamics}, due to the highly influential paper \cite{BYRNESC91}.
 
 Each virtual constraint imposes a relation between joint variables, and by differentiation with respect to time, a relation between joint velocities. As a consequence, for the virtual constraints studied in this paper, the dimension of the zero dynamics is the number of states in the robot's (pinned) model minus twice the number of virtual constraints (which can be at most the number of independent actuators). As explained in \cite{grizzle2017virtual}, the computation of the motor torques to impose virtual constraints parallels the Jacobian computations for the ground reaction forces in a pinned model.
 

\subsection{Allowing Non-holonomic, Time-varying Virtual Constraints} 
\label{sec:VC4ZD}

In this section, we choose $L$ as one of the states of the zero dynamics. 
So that fully actuated and underactuated biped models can be addressed simultaneously, we suppose that the torque distribution matrix $B(q)$ in \eqref{eq:SwingPhasePinnedModel} can be split so that
\begin{equation}
    B(q)u =:  B_a(q)u_a +  B_b(q)u_b,
\end{equation}
where 
$u_a$ is the torque affecting the stance ankle as in Sect.~\ref{sec:LowDimesionDynamics} and $u_b \in \mathbb{R}^{n}$ are actuators affecting the body coordinates, $q_b$. When the robot is underactuated, $B_a(q)$ is an empty column vector. 

We define $n$ virtual constraints as an output zeroing problem of the form
\begin{equation}
 \label{eq:VCeqn}
 y = h(q,L,\tau) = h_0(q) - h_d(x_c,L,\tau)
\end{equation}
where $\tau$ captures time dependence. As in Sect.~\ref{sec:LowDimesionDynamics}, we use $L$ instead of other functions of $\dot{q}$ because $L$ has relative degree three with respect to all actuators except stance ankle torque, while $\dot{q}$ has relative degree one. Hence, the relative degree of $y$ is determined by $q$ once $u_a$ is fixed. Indeed, while $u_b$ is used for imposing the virtual constraints, $u_a$ can be used for shaping the evolution of $x_c$ and $L$ directly. We assume a feedback law, $u_a$, of the form
\begin{equation}
    u_a = \alpha(x_c,L,\tau)
    \label{eq:ua_feedback}
\end{equation}
and note that $u_a$ should respect relevant ankle torque limits and ZMP constraints when $y \equiv 0$.

Following \cite{BYRNESC91,ISI95,WGCCM07}, we make the following specific \textit{regularity assumptions} for the virtual constraints: \newline
\noindent \textbf{A1:} $h$ is at least twice continuously differentiable and $u_a$ is at least once differentiable. \newline
\noindent \textbf{A2:} The virtual constraints \eqref{eq:VCeqn} are designed to identically vanish on a desired nominal solution (gait) $(\bar{q}(t),\dot{\bar{q}}(t), \bar{u}(t))$ of the dynamical model \eqref{eqn:grizzle:modeling:full_hybrid_model_walking} with $\tau(t)=t$, where the solution meets relevant constraints on motor torque, motor power, ground reaction forces, and work space. To be clear, $y$ vanishing means
\begin{equation}
 \label{eq:VCeqn04}
h_0(\bar{q}) - h_d(\bar{x}_c,\bar{L},\tau) \equiv 0
\end{equation}
for $0 \le t\le T$, where $\bar{L}(t)$ is the angular momentum about the contact point, evaluated along the trajectory. \newline
\noindent \textbf{A3:} The \textit{decoupling matrix}
\begin{equation}
 \label{eq:VCeqn03}
 A(q):= \frac{\partial h(q,L,t)}{\partial q} D^{-1}(q)B_b(q) 
\end{equation}
is square and invertible along the nominal trajectory, so that, from \cite{ISI95} and \cite{sadeghian2017passivity}, by treating $u_a$ as a known signal, there exists a feedback controller of the form 
\begin{equation}
    u_b=:\gamma(q, \dot{q}, \tau) +\gamma_a(q, L, \tau)u_a
\end{equation}
resulting in the closed-loop dynamics
\begin{equation}
    \label{eq:IOcontroller01}
    \ddot{y} + K_d \dot{y} + K_p y = 0,
\end{equation}
with $K_d>0$ and $K_p>0$ positive definite. \newline
\noindent \textbf{A4:}  The function
\begin{equation}
    \label{eq:VCeqnLIP01}
    \begin{bmatrix}
y \\ \dot{y} \\ x_c \\ L
 \end{bmatrix}
 \end{equation}
 is full rank and injective in an open neighborhood of the nominal solution $(\bar{q}(t),\dot{\bar{q}}(t))$ $\forall t $. 

From \cite{BYRNESC91,ISI95,WGCCM07}, the above assumptions imply that $(y, \dot{y}, x_c, L)$ is a valid set of coordinates for the full-order swing phase model \eqref{eqn:grizzle:modeling:model_walk}. In particular, 
\begin{enumerate}
    \item there exists an invertible differentiable function $\Phi$ such that
    \begin{equation}
    \label{eq:ZDcoordinates}
        \begin{bmatrix}
        q \\ \dot{q} \\ \tau
        \end{bmatrix} =    \Phi(y, \dot{y}, x_c, L, \tau), \text{~and}
    \end{equation}
    \item the swing phase zero dynamics, that is, the dynamics of the robot compatible with $y \equiv 0$, exists and can be parameterized by $(x_c, L)$.  
\end{enumerate}

\subsection{Zero Dynamics and Approximate Zero Dynamics}
\label{sec:ZeroDynamics}

From Assumptions A1-A4, it follows that the swing phase zero dynamics exists and for $\xi = (x_c, L, \tau)$ can be expressed as 
\begin{equation}
    \label{eqn:ZDGeneral}
    \dot{\xi} = f_{\rm zero}(\xi),
\end{equation}
when $y\equiv 0$. As with the popular pendulum models, the dimension of \eqref{eqn:ZDGeneral} is low, it has two states plus time. Different that the pendulum models, \eqref{eqn:ZDGeneral} is exact. Moreover, tools are known for relating periodic orbits of the hybrid version of \eqref{eqn:ZDGeneral} to corresponding orbits in the full-order model \eqref{eqn:grizzle:modeling:full_hybrid_model_walking}, including their stability properties; see Sect.~\ref{sec:HZDandStability}.

On the basis of \eqref{eqn:ZDnaturalCoordinates} evaluated at \eqref{eq:ZDcoordinates}, the zero dynamics \eqref{eqn:ZDGeneral} can be written more explicitly as
\begin{equation}
    \label{eqn:ZDnaturalCoordinatesZD}
    \begin{aligned}
       \dot{x}_c&= \frac{L}{m z_c(x_c, L, \tau)} + \frac{\dot{z}_c(x_c, L, \tau)}{z_c(x_c, L, \tau)}x_c - \frac{L_c(x_c, L, \tau)}{m z_c(x_c, L, \tau)}\\
       \dot{L}&=m g x_c + u_a(x_c, L, \tau)\\
       \dot{\tau}&=1.
    \end{aligned}
\end{equation}
The state $\dot{\tau}=1$ is included in \eqref{eqn:ZDnaturalCoordinatesZD} because, at hybrid transitions, $\tau$ is reset to zero, that is, $\tau^+:=0$. As discussed above, this reduced-order model is exact along all trajectories of the full-order model for which $y\equiv 0$. 

If one of the virtual constraints in \eqref{eq:VCeqn} is $z_c - H$, that is, the center of mass height is regulated to a constant, then the zero dynamics (exactly) simplifies to 
\begin{equation}
    \label{eqn:ZDnaturalCoordinatesZDConstantH}
    \begin{aligned}
       \dot{x}_c&= \frac{L}{m H}  - \frac{L_c(x_c, L, \tau)}{m H}\\
       \dot{L}&=m g x_c + u_a(x_c, L, \tau)\\
       \dot{\tau}&=1.
    \end{aligned}
\end{equation}
This model is nonlinear and time-varying through $L_c$ and possibly, the feedback control policy chosen for the stance ankle torque, $u_a$. We've argued in Sec.~\ref{sec:PendulumModels} that $L_c$ can be dropped from the model. Doing so results in the ALIP model, \eqref{eqn:JWG:ZD_ALIP}. Hence, the ALIP model is an \textit{approximate swing phase zero dynamics} when the center of mass height is controlled to a constant.

\subsection{Consequences for Closed-loop Stability of the Full-order Model}
\label{sec:HZDandStability}

When the foot placement policy \eqref{eq:Desired_FootplacementAsymptotic} is applied to \eqref{eqn:ZDnaturalCoordinatesZDConstantH} with the rest map \eqref{eq:ALIPresetMAP}, the resulting closed-loop system is a (small) perturbation of a hybrid system that possesses a family of exponentially stable periodic orbits parameterized by $L^{\rm des}$. If the virtual constraints in \eqref{eq:VCeqn} are hybrid\footnote{Hybrid invariance means that if $y$ and $\dot{y}$ are zero before the impact, they will also be zero after the impact. References \cite{MOGR08,ChGrSh09} show how to systematically modify a given set of virtual constraints to achieve hybrid invariance.} invariant for constant $L^{\rm des}$ \cite{WGCCM07,AmGaGrSr2014,yang2021impact}, then 
\begin{itemize}
    \item \eqref{eqn:ZDnaturalCoordinatesZDConstantH} with impact map \eqref{eq:ALIPresetMAP} is the hybrid zero dynamics, and 
    \item an exponentially stable periodic solution of the hybrid zero dynamics is also an exponentially stable solution of the full order closed-loop system for appropriate choices of the feedback gains $K_p$ and $K_d$ in \eqref{eq:IOcontroller01}.
\end{itemize}
Consequently, the closed-loop system would possess a family of exponentially stable periodic orbits parameterized by $L^{\rm des}$. If the virtual constraints are not hybrid invariant, then \eqref{eqn:ZDnaturalCoordinatesZDConstantH} with \eqref{eq:ALIPresetMAP} does not form a hybrid zero dynamics in the sense of \cite{WGCCM07}, but rather a \textit{limit restriction dynamics} \cite[pp. 102]{viola2008control}. Moreover, via the Brouwer Fixed Point Theorem, reference \cite[Theorem 6, pp. 105]{viola2008control} shows each exponentially stable periodic solution of the limit restriction dynamics corresponds to an exponentially stable periodic solution of the full model for appropriate design of the feedback gains in \eqref{eq:IOcontroller01}. 

To illustrate the correspondence between exponentially stable motions of the ALIP and the full-order model, we turn to the Rabbit model controlled via virtual constraints that implement the foot placement control law \eqref{eq:Desired_FootplacementAsymptotic}, the center of mass at a constant height, the torso upright, and adequate foot clearance. We then numerically estimate the Jacobian of the Poincare map for the closed-loop full-order model and compare its dominant eigenvalues to the dominant eigenvalue of the closed-loop ALIP model; see \eqref{eq:PoincareALIP} . 

 In Table \ref{tab:poincare_eigen}, for various values of $\alpha$ in the step placement feedback controller, we show the dominant eigenvalue from the ALIP model and the dominant eigenvalue from the numerically estimated Poincar\'e map. We see that the dominant eigenvalue of the full-order closed-loop system corresponds to the dominant eigenvalue of the ALIP model for $0 \le \alpha \le 0.9$.  The remaining eigenvalues of the full model are (very) small due to the gains chosen in \eqref{eq:IOcontroller01}. In fact, the zero dynamics captures the ``weakly actuated'', slow part of the full-order model that is evolving under the influence of gravity. 

 \begin{table}[h!]
 \center
 \begin{tabular}{ |c|c|c|}
 \hline
 $\alpha$   & ALIP & Rabbit\\
 \hline
 0.9   & 0.81 & 0.781\\
 \hline
 0.8   & 0.64 & 0.601\\
 \hline
 0.7   & 0.49 & 0.442\\
 \hline
  0.6   & 0.36 & 0.299\\
 \hline
  0.5   & 0.25 & 0.168\\
 \hline
  0.4   & 0.16 & 0.052\\
 \hline
  0.3   & 0.09 & 0.013\\
 \hline
  0.2   & 0.04 & 2e-4\\
 \hline
  0.1   & 0.01 & 2e-4 \\
 \hline
  0.0   & 0.00 & 1e-4\\
 \hline
\end{tabular}
\caption{Largest eigenvalues of ALIP and Rabbit under different $\alpha$, for a two-step Poincar\'e map. Because the Poincar\'e map is computed over two steps, the ALIP's largest eigenvalue is $\alpha^2$.}
 \label{tab:poincare_eigen}
 \end{table}
 
 \textbf{Remark:} We numerically obtained the Jacobian of the Poincar\'e map for Rabbit with the foot placement controller by the method symmetric differences; $\delta$ deviations were applied on ten states (Rabbit has 5 degree of freedom) and we measured the corresponding responses after two steps. The $\delta$ was chosen from the set $\{\pm 0.05, \pm 0.1, \pm 0.2, \pm 0.3\}$; see Table \ref{tab:poincare_eigen_delta}.
 
\begin{table}[h!]
 
 \center
 \begin{tabular}{ |c|c|c|c|c|c|c|}
 \hline
 $\alpha$   & ALIP & $\delta$ = $\pm 0.05$ & $\delta$ = $\pm 0.1$ & $\delta$ = $\pm 0.2$ & $\delta$ = $\pm 0.3$\\
 \hline
 0.9   & 0.81 & 0.780 & 0.783& 0.781& 0.779\\
 \hline
 0.8   & 0.64 & 0.602 & 0.602 & 0.601 & 0.599\\
 \hline
 0.7   & 0.49 & 0.442 &0.443 &0.441 &0.439\\
 \hline
  0.6   & 0.36 & 0.299 &0.301 & 0.300 & 0.299\\
 \hline
  0.5   & 0.25 & 0.170 & 0.170 & 0.168 & 0.164\\
 \hline
  0.4   & 0.16 & 0.054 & 0.053 & 0.052 & 0.051\\
 \hline
  0.3   & 0.09 & 0.014 & 0.014 & 0.012 & 0.011\\
 \hline
\end{tabular}
\caption{Numerical support for estimating the Jacobian of the  two-step Poincar\'e map. The dominant eigenvalue of Rabbit model is insensitive to the perturbation used in estimating the Jacobian.}
\label{tab:poincare_eigen_delta}
 \end{table}
 
 \subsection{Non-periodic Walking} 
 
The desired angular momentum, $L^{\rm des}$, determines the fixed point of the Poincar\'e map and hence the walking speed of the robot. While varying $L^{\rm des}$ causes the walking speed to change, the analysis of the controller has only been presented for a constant value of $L^{\rm des}$. Reference \cite{westervelt2003switching} analyzes gait transitions in the formalism of the hybrid zero dynamics when $L^{\rm des}$ is switched ``infrequently'', meaning the closed-loop system is moving from a neighborhood of one periodic orbit to another. References \cite{kolathaya2018input,veer2019input} generalize tools from Input-to-State Stability (ISS) of ODEs to the case of hybrid models. These results apply to time-varying $L^{\rm des}$. The experimental work reported in Sec.~\ref{Sec:ExperimentResults} includes examples of rapidly varying $L^{\rm des}$, turn direction, and ground height. 
 

\subsection{Varying Center of Mass Height}
We have seen that the difference between the zero dynamics of a real robot and a pendulum model is the term related to $L_c$. In previous sections, we have shown that the $L_c$ term has very little effect on the $L$ dynamics when $z_c$ is constant. This observation can be extended to the case when $z_c$ is not constant but virtually constrained by $(x_c, L, \tau)$.

\begin{figure}
    \centering
    \includegraphics[width=0.24\textwidth]{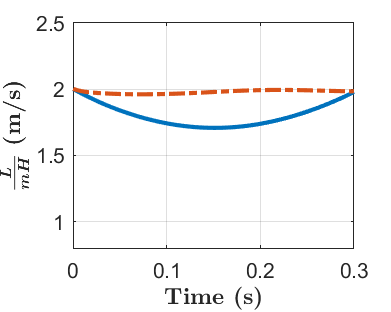}
    \caption{Trajectory of $L$ and its prediction in a  simulation of Rabbit. The instantaneous values are shown in \textcolor{blue}{\bf blue} and the predicted value at the end of the step is shown in \textcolor{red}{\bf red}. In the prediction of $L$, the virtual constraint on center of mass height for the model model and for Rabbit are set to $z_c = 0.6 + 0.05\sin(\frac{T}{2\pi}\tau - \frac{\pi}{2}) + 0.05$, where $T$ is the step time. Large $z_c$ oscillations often occur in running. Here, we modify the ground model to pin the stance foot to the ground, so that we can impose a non-trivial $z_c$ oscillation in periodic walking.}
    \label{fig:varying_height}
\end{figure}

In Fig.~\ref{fig:varying_height}, we illustrate that when the $z_c$ is a function of time, the pendulum dynamics can still be used to predict accurately the zero dynamics of Rabbit. 

\section{Integrating Virtual Constraints and Angular-Momentum-based Foot Placement} \label{sec:virtual_constraint_Cassie}

In this section we generate virtual constraints for a 3D robot such as Cassie. As in \cite{Yukai2018}, we leave the stance toe passive. Consequently, there are nine (9) control variables, listed below from the top of the robot to the end of the swing leg,
\begin{equation} \label{eq:control variables}
\small
h_0=
\begin{bmatrix}
\rm torso \; pitch \\
\rm torso \; roll \\
\rm stance \; hip \; yaw \\
\rm swing \; hip \; yaw \\
p^z_{\rm st \to CoM} \\
p^x_{\rm sw \to CoM} \\
p^y_{\rm sw \to CoM} \\
p^z_{\rm sw \to CoM} \\
\rm swing \;toe \;absolute \;pitch \\
\end{bmatrix}.
\normalsize
\end{equation}
For later use, we denote the value of $h_0$ at the beginning of the current step by $h_0(T^+_{k-1})$. When referring to individual components, we'll use $h_{03}(T^+_{k-1})$, for example.

We first discuss variables that are constant. The reference values for torso pitch, torso roll, and swing toe absolute pitch are constant and zero, while the reference for $p^z_{\rm st \to CoM}$, which sets the height of the CoM with respect to the ground, is constant and equal to $H$.

We next introduce a phase variable
\begin{equation} \label{eq:phase_variable}
    s := \frac{t - T_{k-1}^+}{T}
\end{equation}
that will be used to define quantities that vary throughout the step to create ``leg pumping'' and ``leg swinging''. The reference trajectories of $p^x_{\rm sw \to CoM}$ and $p^y_{\rm sw \to CoM}$ are defined such that:
\begin{itemize}
    \item at the beginning of a step, their reference value is their actual position;
    \item the reference value at the end of the step implements the foot placement strategy in \eqref{eq:Desired_Footplacement}; and
    \item in between a half-period cosine curve is used to connect them, which is similar to the trajectory of an ordinary (non-inverted) pendulum.
\end{itemize}
The reference trajectory of $p^z_{\rm sw \to CoM}$ assumes the ground is flat and the control is perfect: 
\begin{itemize}
    \item at mid stance, the height of the foot above the ground is given by $z_{CL}$, for the desired vertical clearance.
\end{itemize}

The reference trajectories for the stance hip and swing hip yaw angles are simple straight lines connecting their initial actual position and their desired final positions. For walking in a straight line, the desired final position is zero. To include turning, the final value has to be adjusted.  Suppose that a turn angle of $\Delta D^{\rm des}_k$ radians is desired. One half of this value is given to each yaw joint:
\begin{itemize}
\item $+ \frac{1}{2}\Delta D^{\rm des}_k \to \rm swing \; hip \; yaw$; and
\item $- \frac{1}{2}\Delta D^{\rm des}_k \to \rm stance \; hip \; yaw$
\end{itemize}
The signs may vary with the convention used on other robots.

The final result for Cassie Blue is
\begin{equation} \label{eq:control variables reference} \small
\begin{aligned}
&h_{\rm d}(s):=\\
&
\begin{bmatrix}
0 \\
0 \\
(1-s)h_{03}(T_{k-1}^+) + s(-\frac{1}{2}(\Delta D_{k})) \\
(1-s)h_{04}(T_{k-1}^+) + s(\frac{1}{2}(\Delta D_{k})) \\
H \\
\frac{1}{2}\big[(1+\cos(\pi s))h_{06}(T_{k-1}^+) + (1 - \cos(\pi s))p^{x ~ \rm des}_{\rm sw \to CoM}(T_k^-)\big] \\
\frac{1}{2}\big[(1+\cos(\pi s))h_{07}(T_{k-1}^+) + (1 - \cos(\pi s))p^{y ~ \rm des}_{\rm sw \to CoM}(T_k^-)\big] \\
4 z_{cl} (s-0.5)^2+(H-z_{CL}); \\
0 \\
\end{bmatrix}
\end{aligned}.
\end{equation}
When implemented with an Input-Output Linearizing Controller\footnote{The required kinematic and dynamics functions are generated with FROST \cite{frost}.} so that $h_0$ tracks $h_d$, the above control policy allows Cassie to move in 3D in simulation.

\section{Practical Implementation on Cassie} \label{Sec:Experiment}
\begin{figure}
    \centering
    \includegraphics[width=0.45\textwidth]{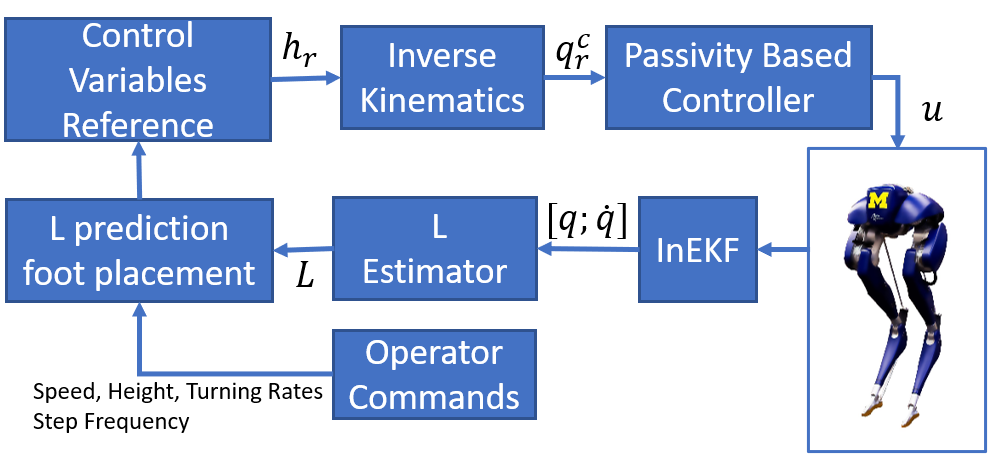}
    \caption{Block diagram of the implemented controller.}
    \label{fig:Control_Diagram}
    \vspace*{-.5cm}
\end{figure}
This section resolves several issues that prevent the basic controller from being implemented on Cassie Blue.

\subsection{IMU and EKF}
In a real robot, an IMU and an EKF are needed to estimate the linear position and rotation matrix at a fixed point on the robot, along with their derivatives. Cassie uses a VectorNav IMU. We used the Contact-aided Invariant EKF developed in \cite{hartley2020contact,invariant-ekf} to estimate the torso velocity. With these signals in hand, we could estimate angular momentum about the contact point.

\subsection{Filter for Angular Momentum} \label{subsec:AMKF}
Angular Momentum about the contact toe could be computed directly from estimated $[q,\dot{q}]$, but it is noisy. We used a Kalman Filter to improve the estimation. The models we used are

\begin{equation}
    \begin{aligned}
  \text{Prediction:    } & L^y(k) = AL^y(k-1) + Bu(k) + \delta\\
  \\
  \text{Correction:  }& L^y_{\rm \;obs}(k) = C L^y(k) + \epsilon
  \end{aligned}
\end{equation}
where $A = B = C = 1$, $u(k) = (mgx_{\rm c}(k) + u_a(k))\Delta T$.
\chg{
The update formula for angular momentum is 
\begin{equation}
    L^y(k) = (I-K(k)C)(AL^y(k-1) + Bu(k)) + K(k) L^y_{\rm \;obs}(k)
\end{equation}
The Kalman Gain $K(k)$ is obtained following the algorithm described in \cite[Sec 3.2]{thrun2002probabilistic}.}


\subsection{Inverse Kinematics}
Input-Output Linearization does not work well in experiments\cite{Yukai2018,CLFKoushil,WEBUGR04}. To use a passivity-based controller for tracking that is inspired by \cite{sadeghian2017passivity}, we need to convert the reference trajectories for the variables in \eqref{eq:control variables} to reference trajectories for Cassie's actuated joints,
\begin{equation}
    q^{\rm act} = 
    \begin{bmatrix}
    \rm torso \; pitch \\
    \rm torso \; roll \\
    \rm stance \; hip \; yaw \\
    \rm swing \; hip \; yaw \\
    \rm stance \; knee \; pitch \\
    \rm swing \; hip \; roll \\
    \rm swing \; hip \; pitch \\
    \rm swing \; knee \; pitch \\
    \rm swing \; toe \; pitch \\
    \end{bmatrix}.
\end{equation}
Iterative inverse kinematics is used to convert the controlled variables in \eqref{eq:control variables} to the actuated joints. 

\subsection{Passivity-based Controller}

We adapt the passivity-based controller developed in \cite{sadeghian2017passivity} to achieve joint-level tracking. This method takes the full-order model of robot into consideration and, on a perfect model, will drive the virtual constraints asymptotically to zero. The derivation is given in the Appendix. 

\subsection{Springs}
On the swing leg, the spring deflection is small and thus we are able to assume the leg to be rigid. On the stance leg, the spring deflection is non-negligible and hence requires compensation. While there are encoders on both sides of the spring to measure its deflection, direct use of this leads to oscillations. The deflection of the spring is instead estimated through a simplified model.  

\subsection{COM Velocity in the Vertical Direction}

When Cassie's walking speed exceeds one meter per second, the assumption that $v^z_{\rm CoM} \approx 0$ breaks down due to spring and imperfect low level control, and \eqref{eq:AM_impact_relation_1} is no longer valid. 
Hence, we use
\begin{equation} \label{eq:AM_impact_relation_2} \small
    L^y(T_k^+) = L^y(T_k^-) + mv^z_{\rm c}(T_k^-)(p^x_{\rm sw \to CoM}(T_k^-) - p^x_{\rm st \to CoM}(T_k^-)).
\end{equation}
From this, the foot placement is updated to
\begin{multline} \label{eq:foot_placement_2} \small
    p^{x \; \rm des}_{\rm sw \to CoM}(T_k^-) =  \frac{ L^{y \; \rm des}(T_{k+1}^-)}{ m(H\ell \sinh(\ell T) - v^z_{\rm CoM})\cosh(\ell T) } - \\
    \frac {(L^y(T_k^-)+mv^z_{\rm CoM}(T_k^-)p^x_{\rm st \to CoM}(T_k^-))\cosh(\ell T) }{m(H\ell \sinh(\ell T) - v^z_{\rm CoM})\cosh(\ell T)}.
\end{multline}
\chg{ \textbf{Remark:} In our experiments $v^z_{\rm CoM}$ becomes negative at the end of a step when the robot is walking fast. If we still use \eqref{eq:Desired_Footplacement} to decide foot placement, which is based on the reset map \eqref{eq:AM_impact_relation_1}, in the lateral direction $L^x(T_k^+)$ will be overestimated. This in turn leads to the lateral foot placement being commanded further from the body than it should be. At the end of the next step, the magnitude of $L^x$ will be larger than expected, requires even further lateral foot placement from the body. The final phenomenon is abnormally large step width.}




\section{Experimental Results} \label{Sec:ExperimentResults}
The controller was implemented on Cassie Blue. The closed-loop system consisting of robot and controller was evaluated in  a number of situations that are itemized below.
\begin{itemize}
    \item \textbf{Walking in a straight line on flat ground.} Cassie could walk in place and walk stably for speeds ranging from zero to $2.1$ m/s. 
    
    \item \textbf{Diagonal Walking.} Cassie is able to walk simultaneously forward and sideways on grass, at roughly $1$ m/s in each direction.
    
    \item \textbf{Sharp turn.} While walking at roughly $1$ m/s, Cassie Blue effected a $90^o$ turn in six steps, without slowing down. 
    
    \item \textbf{Rejecting the classical kick to the base of the hips.} Cassie was able to remain upright under ``moderate'' kicks in the longitudinal direction. The disturbance rejection in the lateral direction is not as robust as the longitudinal, which is mainly caused by Cassie's physical design: small hip roll motor position limits. 
    
    \item \textbf{Finally we address walking on rough ground}. Cassie Blue was tested on the iconic Wave Field of the University of Michigan North Campus. The foot clearance was increased from 10 cm to 20 cm to handle the highly undulating terrain. Cassie is able to walk through the``valley'' between the large humps with ease at a walking pace of roughly $0.75$ m/s, without falling in all tests. The row of ridges running east to west in the Wave Field are roughly 60 cm high, with a sinusoidal structure. We estimate the maximum slope to be 40 degrees. Cassie is able to cross several of the large humps in a row, but also fell multiple times. On a more gentle, straight grassy slope of roughly 22 degrees near the laboratory, Cassie can walk up it with no difficulty with 20cm foot clearance. 
\end{itemize} 

\begin{figure}[ht!]
    \centering
    \begin{subfigure}{.45\textwidth}
      \centering
      \includegraphics[width=\textwidth]{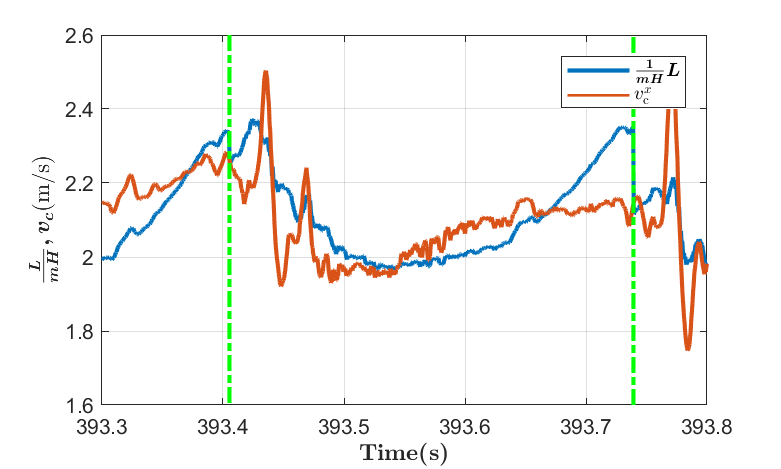}
      \caption{$L$ and $v_c$}
    \end{subfigure}
    \begin{subfigure}{.45\textwidth}
      \centering
      \includegraphics[width=\textwidth]{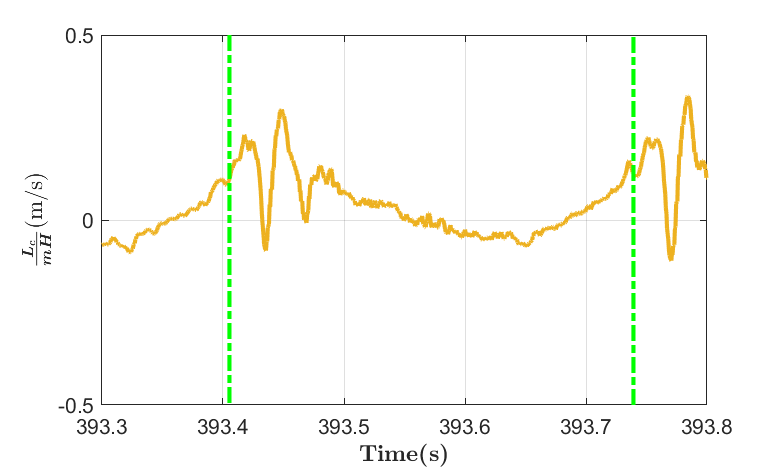}
      \caption{\chg{$L_c$}}
    \end{subfigure}
        \begin{subfigure}{.45\textwidth}
      \centering
      \includegraphics[width=\textwidth]{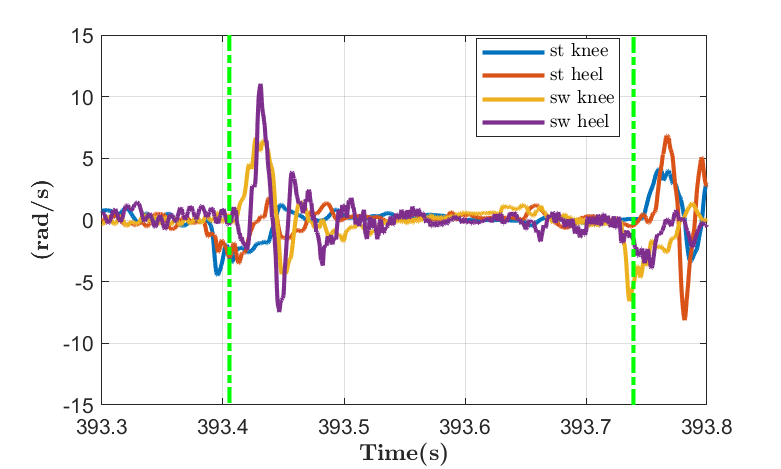}
      \caption{\chg{Spring deflection rates for stance and swing legs}}
    \end{subfigure}
    \caption{Experimental data from Cassie walking forward at about 2m/s. \chg{To ensure a fair comparison, $L$ is not smoothed by Kalman Filter described in \ref{subsec:AMKF}. $L$, $v_c$ and $L_c$ are computed from the same states $[q,\dot{q}]$. \textcolor{red}{$v^x_c$} and \textcolor{Goldenrod}{$L_c$} oscillates at the beginning of a step because of their relative degree one nature, in particular, they are heavily affected by the spring oscillation just after impact.} \textcolor{blue}{$L$} is mostly smooth because it has relative degree three, except near impact when the robot is in double support phase and $L$ has relative degree one. The sudden jump in $L$ at impact is caused by nonzero $v^z_c$, \chg{The smoothness difference shows another advantage of \textcolor{blue}{$L$}: it can be used in feedback control without being heavily filtered.}}
    \label{fig:L_vc_exp}
\end{figure}

\begin{figure}[ht!]
    \centering
    \includegraphics[width=0.4\textwidth]{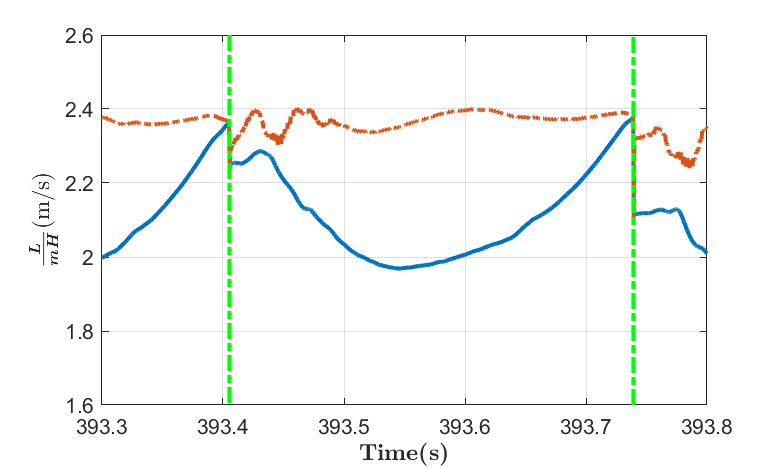}
    \caption{\chg{Prediction made in experiment from Cassie walking forward at about 2m/s. The instantaneous values are shown in \textcolor{blue}{\bf blue} and the predicted value at the end of the step is shown in \textcolor{red}{\bf red}.  The Kalman Filter described in \ref{subsec:AMKF} has been applied}.}
    \label{fig:L_pred_exp}
\end{figure}

\chg{The experimental data is analyzed in Fig. \ref{fig:L_vc_exp} and \ref{fig:L_pred_exp}. The figures support the advantages of using $L$ to indicate robot status and the accuracy of ALIP model, as discussed in Sec \ref{sec:AngularMomentumGeneral} and \ref{sec:PendulumModels}.}

\begin{figure}[ht!]
\begin{subfigure}{.242\textwidth}
  \centering
  \includegraphics[width=\textwidth]{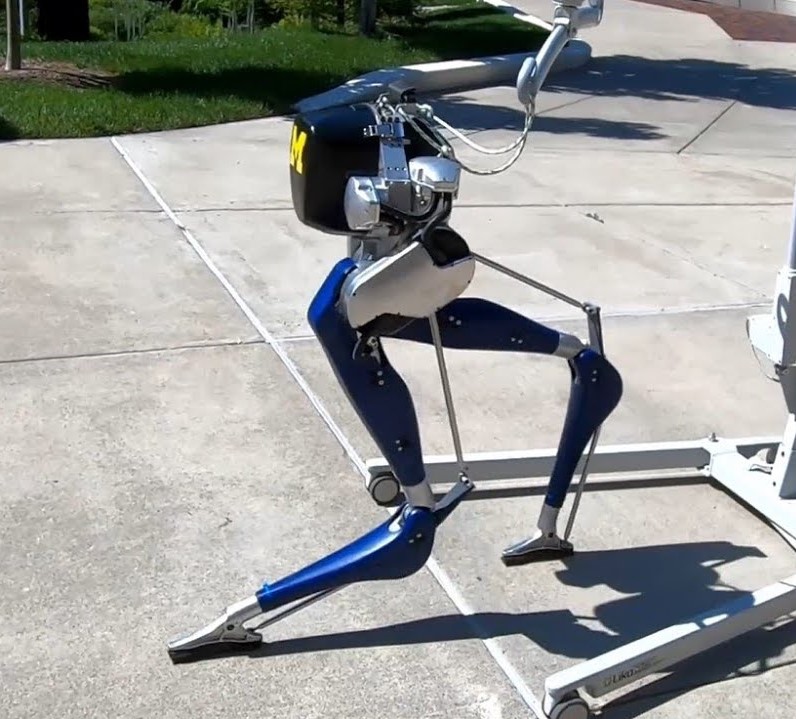}
  \caption{Fast Walking}
  \label{fig:FastWalking}
\end{subfigure}
\begin{subfigure}{.238\textwidth}
  \centering
  \includegraphics[width=\textwidth]{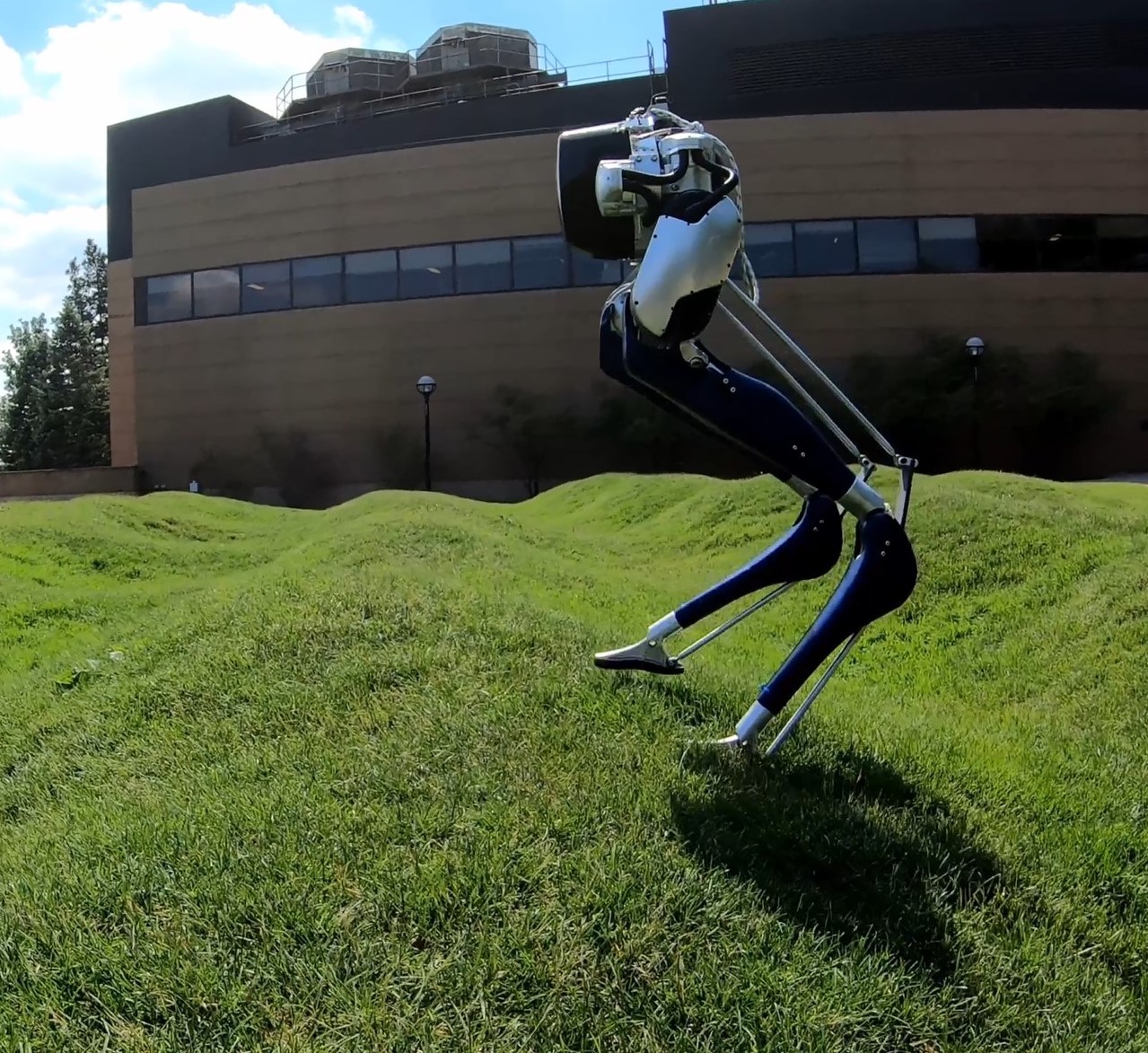}
  \caption{Rough Terrain}
  \label{fig:RoughTerrain}
\end{subfigure}
\begin{subfigure}{.488\textwidth}
  \centering
  \includegraphics[width=\textwidth]{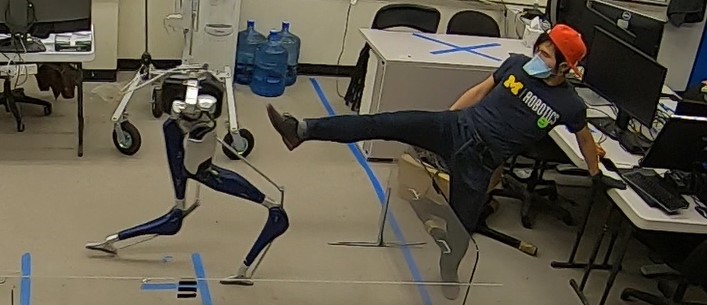}
  \caption{Disturbance Rejection}
  \label{fig:KickTest}
\end{subfigure}
\begin{subfigure}{.488\textwidth}
  \centering
  \includegraphics[width=\textwidth]{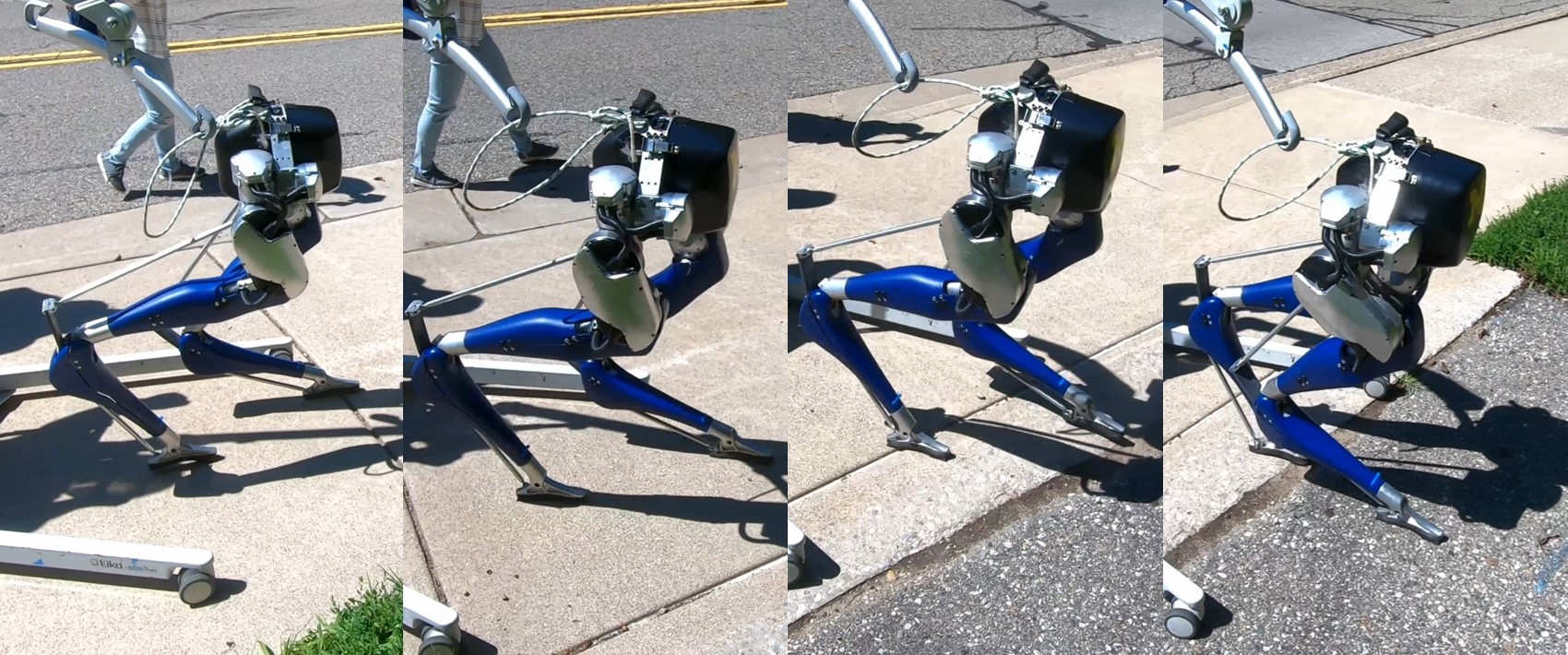}
  \caption{A Fast 90 Degree Turn with a Long Stride}
  \label{fig:Turning}
\end{subfigure}
\caption{Images from several closed-loop experiments conducted with Cassie Blue and the controller developed in this paper. A short video compilation of these experiments is available in \cite{AMLIP_Video}. Longer versions can be found in \cite{LabYoutube}.}
\label{fig:ExperimentResult}
\vspace*{-.5cm}
\end{figure}

\section{Conclusions} \label{Sec:Conclusion}
We established connections between various approximate pendulum models that are commonly  used for heuristic controller design and those that are more common in the feedback control literature where formal stability guarantees are the norm. \chg{The paper clarified commonalities and differences in the two perspectives for using low-dimensional models.} In the process of doing so, we argued that models based on angular momentum about the contact point provide more accurate representations of robot state than models based on linear velocity. Specifically, we showed that an approximate (pendulum or zero dynamics) model parameterized by angular momentum provides better predictions for foot placement on a physical robot (e.g., legs with mass) than does a related approximate model parameterized in terms of linear velocity. \chg{While ankle torque, if available, can achieve small changes in angular momentum during a step, the limitations of foot roll prevent ankle torque from achieving large changes in angular momentum during a step. For this reason, we focused our analysis on regulating  angular momentum about the contact point step-to-step and not within a step.} We implemented a one-step-ahead angular-momentum-based controller on Cassie, a 3D robot, and demonstrated high agility and robustness in experiments. Using our new controller, Cassie was able to accomplish a wide range of tasks with nothing more than common sense task-based tuning: a higher step frequency to walk at 2.1 m/s and extra foot clearance to walk over slopes exceeding 22 degrees. Moreover, in the current implementation, there is no optimization of trajectories used in the implementation on Cassie. The robot's performance is currently limited by the hand-designed trajectories leading to joint-limit violations and foot slippage. These limitations will be mitigated by incorporating optimization and or constraints via MPC.

\chg{In addition to the foot placement scheme described in this paper, the ALIP model can be combined with many other control methods that have been implemented on the LIP model: it can be used with ZMP in an MPC scheme as in \cite{KaKaKaFuHaYoHi03,nishiwaki2006high, wieber2006trajectory}, or used with Capture Point as in \cite{pratt2006capture,englsberger2011bipedal}, or used with an estimated $L_c$ \cite{seyde2018inclusion,lee2007reaction}. As has already been done in  \cite{powell2016mechanics}, $L$ can be regulated by controlling the center of mass velocity just before impact, as well.}


\vspace{-2mm}
\appendix

\appendices

\subsection{Constant Pendulum Length}
\label{sec:ThetacModel}

Suppose that one component of the virtual constraints in \eqref{eq:VCeqn} is $r_c(q) - R $, where $R$ is a constant. Then $y\equiv 0$  yields $r_c=R$, simplifying \eqref{eqn:JWG:ZD_theta_c} to
        \begin{equation}
    \label{eqn:JWG:ZD_theta_cRconstant}
       \begin{aligned}
       \dot{\theta}_c&= \frac{L-L_c}{m R^2}  \\
       \dot{L}&=m g R \sin(\theta_c) + u_a.
    \end{aligned}
\end{equation}
 At this point, no approximations have been made and the models is valid everywhere that $r_c(q) \equiv R$. An interesting aspect of this pendulum model is that it does not depend on $\dot{R}$, and thus imperfections in a achieving the virtual constraint $r_c=R$ have a smaller effect here than in \eqref{eqn:ZDnaturalCoordinatesConstantHeight}, where $\dot{z}$ would appear when $z_c \neq H$, or in \eqref{eqn:zdOttConstantHeight}, where both $\dot{z}_c$ and $\ddot{z}_c$ would appear. 
 
 As with \eqref{eqn:ZDnaturalCoordinatesConstantHeight}, the model \eqref{eqn:JWG:ZD_theta_cRconstant} is driven by the strongly actuated states $q_b, \dot{q_b}$ through $L_c$ and the same discussion applies. Dropping $L_c$ in \eqref{eqn:ZDnaturalCoordinates} results in 
\begin{equation}
    \label{eqn:JWG:ZD_theta_cRconstantDropLc}
       \begin{aligned}
       \dot{\theta}_c&= \frac{L}{m R^2}  \\
       \dot{L}&=m g R \sin(\theta_c) + u_a,
    \end{aligned}
\end{equation}
which is nonlinear in $\theta_c$. However, for $R=1$ and a step length of $60$~cm, $\max{\theta_c} \approx \pi/6$, and for $70$~cm, $\max{\theta_c} \approx \pi/4$, giving simple bounds on the approximation error,
$$
\begin{aligned}
  \frac{1}{\pi/6}\int_0^{\pi/6} (\theta - \sin(\theta))  d\theta &< 0.006 \\
  \frac{1}{\pi/4}\int_0^{\pi/4} (\theta - \sin(\theta))  d\theta &< 0.02.
\end{aligned}
$$
Moreover, if desired, one can chose $K$ to set
$$
  \frac{1}{\theta_{\rm max}}\left| \int_0^{\theta_{\rm max}} (K \theta - \sin(\theta))  d\theta \right| =0.
$$
For $\theta_{\rm max} = \pi/4$, the value is $K\approx 0.95$.
While a linear approximation is useful for having a closed-form solution, numerically integrating the nonlinear model \eqref{eqn:JWG:ZD_theta_cRconstantDropLc} in real time is certainly feasible. 

The discussion on the approximate zero dynamics can be repeated here. The associated impact map is nonlinear and can be linearized about a nominal solution.

\subsection{Passivity-based Input-Output Stabilization}

The material presented here adapts the original work in \cite{sadeghian2017passivity} to a floating-base model
\begin{equation}
\label{eq:FloatingBaseModel}
D(q)\ddot{q} + H(q,\dot{q}) = Bu + J_{\rm s}^\top \tau_{\rm s}+ J_{\rm g}(q)^\top \tau_{\rm g},
\end{equation}
with $u$ the vector of motor torques, $\tau_{\rm s}$ the spring torques, and $\tau_{\rm g}$ is the contact wrench. During the single support phase, the blade-shape foot on Cassie provides five holonomic constraints, leaving only foot roll free. To simplify the problem, we also assume the springs are rigid, adding two constraint on each leg. These constraints leave the original 20-degree-of-freedom floating base model with 11 degrees of freedom.

The constraints mentioned above can be written as
\begin{align}
\begin{cases}
\label{eq:ground_spring_constraints}
&J_{\rm s}\ddot{q} = 0 \\
&J_{\rm g}(q)\ddot{q} + \dot{J}_{\rm g}(q)\dot{q}  = 0.
\end{cases}
\end{align}
Combining \eqref{eq:FloatingBaseModel} and \eqref{eq:ground_spring_constraints} yields the full model for Cassie in single support,
\begin{equation}
\underbrace{
\label{eq:ExtendedModel}
\begin{bmatrix}
D  &  -J_{\rm s}^\top & -J_{\rm g}^\top \\
J_{\rm s}  & 0 & 0 \\
J_{\rm g} & 0 & 0
\end{bmatrix}
}_{\widetilde{D}}
\underbrace{
\begin{bmatrix}
\ddot{q} \\
\tau_{\rm s} \\
\tau_{\rm g}
\end{bmatrix}
}_{f}
+
\underbrace{
\begin{bmatrix}
H\\
0\\
\dot{J}_{\rm g}\dot{q}
\end{bmatrix}
}_{\widetilde{H}}
=
\underbrace{
\begin{bmatrix}
B \\
0 \\
0 
\end{bmatrix}
}_{\widetilde{B}}
u
\end{equation}
For simplicity we assume that the components of $q$ have already been ordered such that $q = [q_c,q_u]^T$, where $q_c$ are the coordinates chosen to be controlled and $q_u$ are the free coordinates. Define $\lambda = [q_u,\tau_s,\tau_g]^T$ and partition \eqref{eq:ExtendedModel} as
\begin{equation}
\begin{cases}
\widetilde{D}_{11}\ddot{q}_c + \widetilde{D}_{12}\lambda + \widetilde{H}_1 = \widetilde{B}_1u \\
\widetilde{D}_{21}\ddot{q}_c + \widetilde{D}_{22}\lambda + \widetilde{H}_2 = \widetilde{B}_2u \\
\end{cases}
\end{equation}
The vector $\lambda$ can be eliminated from these equations, resulting in
\begin{align}
\label{eq:controlledjointEOM}
\bar{D}\ddot{q}_c + \bar{H}&= \bar{B}u,
\end{align}
where
\begin{align*}
    \bar{D} &= \widetilde{D}_{11}-\widetilde{D}_{12}\widetilde{D}_{22}^{-1}\widetilde{D}_{21} \\
    \bar{H} &= \widetilde{H}_1 - \widetilde{D}_{12}\widetilde{D}_{22}^{-1}\widetilde{H}_2 \\
    \bar{B} &= \widetilde{B}_1 - \widetilde{D}_{12}\widetilde{D}_{22}^{-1}\widetilde{B}_2.
\end{align*}
We note that $\widetilde{D}_{22}$ being invertible and $\bar{D}$ being positive definite both follow from $\tilde{D}$ being positive definite. Later, we need $\bar{B}$ to be invertible; \textit{this is an assumption} similar to that in \eqref{eq:VCeqn03}. Equation \eqref{eq:controlledjointEOM} is what we will focus on from here on.

For the Passivity-based Controller, the error dynamic for $y := q_c - q_r$ and is designed to be \cite{ott2008cartesian}
\begin{align}
\label{eq:PBCErrorDynamic}
    \bar{D}\ddot{y} + (\bar{C} +k_d)\dot{y} + k_p y = 0,
\end{align}
where $\bar{C}$ is the Coriolis/centrifugal matrix in $\bar{H}$ and it is chosen such that $\dot{\bar{D}}=\bar{C}+\bar{C}^\top$. From \eqref{eq:controlledjointEOM} and  \eqref{eq:PBCErrorDynamic}, we have
\begin{align}
    u = \bar{B}^{-1}(\bar{D}  \ddot{q}_r + \bar{H}) - \bar{B}^{-1}(k_py+ (\bar{C}+k_d)\dot{y}).
\end{align}

Compared with a standard Input-Output Linearization controller, whose error dynamics and command torque are
\begin{gather}
    \ddot{y} + k_d \dot{y} + k_py = 0, ~~\text{and} \\
    u = \bar{B}^{-1}(\bar{D}\ddot{q}_r + \bar{H}) -\bar{B}^{-1}\bar{D} ( k_py+k_d\dot{y}),
\end{gather}
the passivity-based controller induces less cancellation of the robot's dynamics, and if $k_p$ and $k_d$ are chosen to be diagonal matrices, the tracking errors are approximately decoupled because, for Cassie, $\bar{B}^{-1}$ is close to diagonal.
This controller provides improved tracking performance over the straight-up PD implementation in \cite{Yukai2018}.

\section*{Acknowledgment}
Toyota Research Institute provided funds to support this work. Funding for J. Grizzle was in part provided by NSF Award No.~1808051. The
authors thank Omar Harib and Jiunn-Kai Huang for their assistance in the
experiments.


\bibliographystyle{unsrt}
\balance
\bibliography{BibFiles/Bib2020July.bib,BibFiles/bib2.bib}


\end{document}